\documentclass[10pt,twocolumn,letterpaper]{article}

\usepackage[pagenumbers]{cvpr}

\usepackage{makecell}
\usepackage{pifont}
\newcommand{\cmark}{\ding{51}}%
\newcommand{\xmark}{\ding{55}}%
\usepackage{afterpage}
\definecolor{others}{rgb}{0, 0, 0}
\definecolor{barrier}{rgb}{1, 0.47058824, 0.19607843}
\definecolor{bicycle}{rgb}{1, 0.75294118, 0.79607843}
\definecolor{bus}{rgb}{1, 1, 0.0}
\definecolor{car}{rgb}{0.0, 0.58823529, 0.96078431}
\definecolor{construction}{rgb}{0, 1, 1}
\definecolor{motorcycle}{rgb}{0.78431372549 , 0.70588235294, 0}
\definecolor{pedestrian}{rgb}{1, 0, 0}
\definecolor{cone}{rgb}{1, 0.94117647, 0.58823529}
\definecolor{trailer}{rgb}{0.52941176, 0.23529412, 0}
\definecolor{truck}{rgb}{0.62745098, 0.1254902, 0.94117647}
\definecolor{driveable}{rgb}{1, 0, 1}
\definecolor{flat}{rgb}{0.54509804,0.5372549,0.5372549}
\definecolor{sidewalk}{rgb}{0.29411765,0,0.29411765}
\definecolor{terrain}{rgb}{0.58823529,0.94117647,0.31372549}
\definecolor{manmade}{rgb}{0.90196078,0.90196078,0.98039216}
\definecolor{vegetation}{rgb}{0,0.68627451,0}
\definecolor{softgray}{RGB}{245,245,245}

\usepackage{cuted}
\definecolor{cvprblue}{rgb}{0.21,0.49,0.74}
\usepackage[pagebackref,breaklinks,colorlinks,allcolors=cvprblue]{hyperref}

\def\paperID{*****}
\def\confName{CVPR}
\def\confYear{2026}

\def\paperName{ShelfOcc}

\title{\paperName{}: Native 3D Supervision beyond LiDAR for\\Vision-Based Occupancy Estimation}

\author{
\begin{tabular}{c}
    \begin{tabular}{ccc}
        Simon Boeder$^{1,2}$ &
        Fabian Gigengack$^{1}$ &
        Simon Roesler$^{1}$ \\
        {\tt\small simon.boeder@de.bosch.com} &
        {\tt\small fabian.gigengack@de.bosch.com} &
        {\tt\small simon.roesler@de.bosch.com}
    \end{tabular}
    \\[0.7em]
    \begin{tabular}{cc}
        Holger Caesar$^{3}$ &
        Benjamin Risse$^{2}$ \\
        {\tt\small h.Caesar@tudelft.nl} &
        {\tt\small b.risse@uni-muenster.de}
    \end{tabular}
    \\[1.0em]
    $^{1}$Bosch Research \quad
    $^{2}$University of M\"unster \quad
    $^{3}$TU Delft
\end{tabular}
}

\begin{document}
\maketitle

\begin{strip}
    \vspace{-12mm}
    \centering
    \includegraphics[page=1, trim=0cm 1.62cm 0cm 0cm, clip, width=\textwidth]{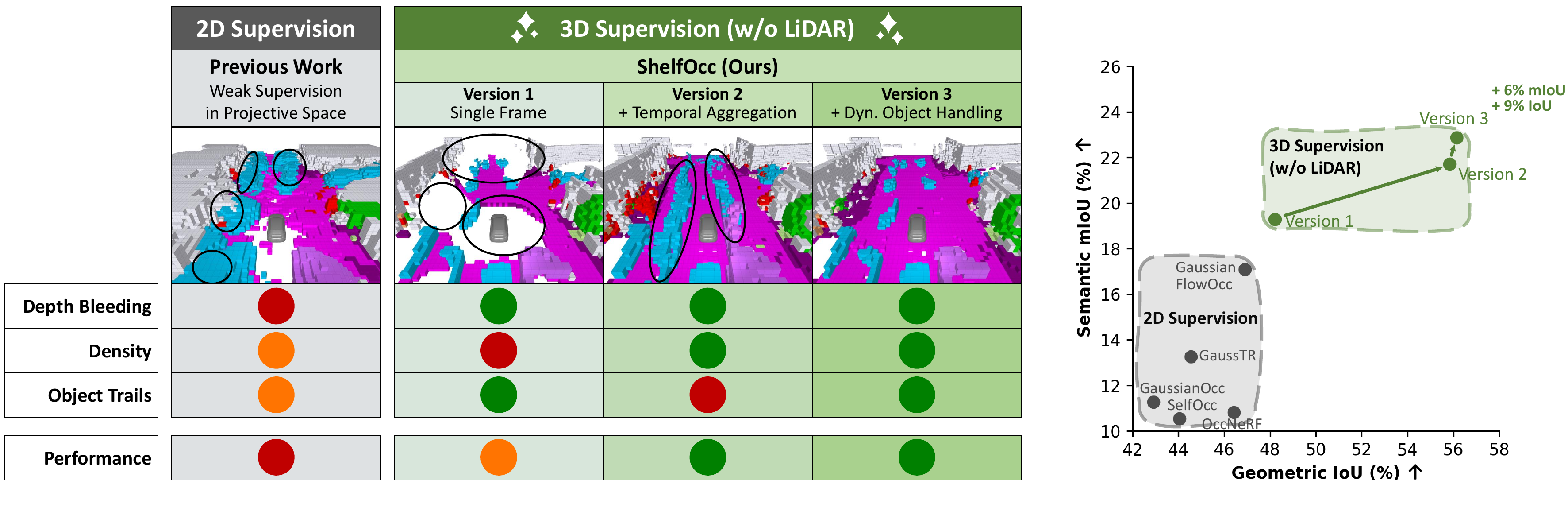}
    \captionof{figure}{
        \textbf{Contributions of \textit{\paperName{}}.}
        We propose a shift in supervision strategy for weakly/shelf-supervised occupancy estimation.
        Unlike prior 2D rendering-based approaches, which are prone to depth bleeding, \textit{\paperName{}} trains occupancy networks directly in native 3D voxel space with pseudo-labels generated using a combination of geometric and semantic FMs.
        By accumulating and filtering static geometry while handling dynamic objects separately, our approach yields clean and consistent 3D supervision relying only on images, without LiDAR.
        This shift in supervision leads to a significant performance gain over previous methods, as illustrated on the right.
    }
    \label{fig:teaser}
    \vspace{-2mm}
\end{strip}

\begin{abstract}
Recent progress in self- and weakly supervised occupancy estimation has largely relied on 2D projection or rendering-based supervision, which suffers from geometric inconsistencies and severe depth bleeding.
We thus introduce \paperName{}, a vision-only method that overcomes these limitations without relying on LiDAR.
\paperName{} brings supervision into native 3D space by generating metrically consistent semantic voxel labels from video, enabling true 3D supervision without any additional sensors or manual 3D annotations.
While recent vision-based 3D geometry foundation models provide a promising source of prior knowledge, they do not work out of the box as a prediction due to sparse or noisy and inconsistent geometry, especially in dynamic driving scenes.
Our method introduces a dedicated framework that mitigates these issues by filtering and accumulating static geometry consistently across frames, handling dynamic content and propagating semantic information into a stable voxel representation.
This data-centric shift in supervision for weakly/shelf-supervised occupancy estimation allows the use of essentially any SOTA occupancy model architecture without relying on LiDAR data.
We argue that such high-quality supervision is essential for robust occupancy learning and constitutes an important complementary avenue to architectural innovation.
On the Occ3D-nuScenes benchmark, \paperName{} substantially outperforms all previous weakly/shelf-supervised methods (up to a $34\%$ relative improvement), establishing a new data-driven direction for LiDAR-free 3D scene understanding.
\end{abstract}

\vspace{-4mm}
\section{Introduction}\label{sec:intro}
Accurate and efficient 3D occupancy estimation is fundamental for safe and reliable autonomous driving, providing a dense understanding of the environment which is crucial for planning and navigation~\cite{shi2024grid, xu2025survey}.
While significant advancements have been made, many state-of-the-art methods still depend heavily on dense 3D ground truth annotations derived from LiDAR sensors.
This dependency is a major bottleneck for scalability and real-world application, as manual dense 3D annotation is extremely costly and labor-intensive~\cite{tempfli2025vespa}.
Moreover, fleet vehicles are rarely equipped with such reference sensors, preventing their data from being utilized for supervised training.

To overcome the challenges of 3D label acquisition, researchers have explored weakly/shelf-supervised approaches, often relying on 2D annotations or photometric losses through differentiable rendering.
We use the term \emph{shelf-supervised}~\cite{9577840, khurana2024shelf} to denote a form of self-supervision that relies on off-the-shelf foundation models as sources of geometric and/or semantic priors.
Methods like SelfOcc~\cite{huang2023selfocc}, OccNeRF~\cite{zhang2023occnerf}, and GaussianFlowOcc~\cite{Boeder_2025_ICCV} utilize techniques such as Neural Radiance Fields (NeRF)~\cite{mildenhall2021nerf} or 3D Gaussian Splatting (3DGS)~\cite{kerbl20233d} to render 3D scene representations back to 2D image space.
This enables supervision from easily obtainable 2D cues, such as semantic segmentation masks produced by models like GroundedSAM~\cite{ren2024grounded} and monocular depth maps from, e.g., Metric3D~\cite{yin2023metric3d}.
However, a critical limitation persists: learning complex 3D geometry solely from 2D image-based losses is inherently difficult.
This frequently results in artifacts such as depth bleeding, where models fail to precisely capture the volumetric extent of objects along viewing rays, since 2D signals primarily provide information about the visible object boundary.
To provide more complete 3D supervision, rendering-based approaches rely on temporal consistency, which requires handling dynamic objects and further complicates the training while reducing but not removing the issue of depth bleeding.
These ambiguities limit performance compared to methods with direct 3D supervision.

The growing availability of off-the-shelf 3D vision foundation models (FMs) has opened new opportunities for leveraging pretrained geometric priors in downstream perception tasks. 
Models such as the Visual Geometry Grounded Transformer (VGGT)~\cite{wang2025vggt} and MapAnything~\cite{keetha2025mapanything} can infer detailed 3D scene attributes from images, making them appealing sources of supervision for 3D occupancy learning.
Trained on vast amounts of 3D-annotated data, these FMs can infer camera parameters, depth maps, and dense 3D point clouds from multiple images in a single forward pass.
However, as illustrated in \cref{fig:teaser}, directly applying such FMs, which typically assume static scenes and consistent camera parameters, to dynamic multi-camera driving sequences poses several challenges.
These include handling non-static elements and integrating semantic information into 3D space.
A naïve frame-wise application (\cref{fig:teaser} Version 1) of FMs produces sparse and incomplete labels, while simple temporal aggregation (\cref{fig:teaser} Version 2) leads to model violations which manifest themselves in dynamic object trails and ghosting artifacts.

In this paper, we propose \textit{\paperName{}}, a shelf-supervised learning framework that leverages 3D geometry FMs to generate high-quality 3D pseudo-labels and train high-performing occupancy networks.
Our proposed 3D pseudo-label generation pipeline separates dynamic and static scene parts, filters noisy and wrong predictions, accumulates the static scene across the sequence and re-introduces dynamic objects per frame.
\paperName{} offers a plug-and-play solution for supervising any occupancy network with 3D labels, solely from camera images.
Our central hypothesis is that direct 3D supervision, even when derived from FM-generated pseudo-labels, can substantially enhance the geometric understanding of occupancy networks, yielding superior performance compared to 2D-supervised counterparts, without requiring costly 3D annotations.
Owing to its modular design, \paperName{} can be seamlessly integrated into existing occupancy prediction pipelines, providing a scalable framework for 3D supervision and enabling LiDAR-free training of state-of-the-art occupancy architectures.
We demonstrate that enhancing the supervision signal can lead to substantial gains in occupancy estimation without modifying the network design itself. 

\noindent In summary our contributions are as follows:
\begin{itemize}
    \item \textbf{Paradigm shift for shelf-supervised occupancy estimation.}
    We introduce \textit{\paperName{}}, a novel framework that enables direct and native 3D shelf-supervision for occupancy estimation by leveraging the emergent capabilities of off-the-shelf geometric and semantic foundation models, eliminating the need for LiDAR, manual 3D annotations or 2D rendering supervision.

    \item \textbf{Vision-only 3D pseudo-label generation.}
    We develop a sophisticated, semantics-aware pipeline that separates static and dynamic scene components, accumulates static geometry across time, reintroduces dynamic objects per frame, and filters low-confidence predictions to produce clean and consistent 3D voxel pseudo-labels.

    \item \textbf{Data-centric performance gains.} 
    We empirically show that our enhanced 3D supervision signal yields substantial performance improvements, surpassing all prior shelf-supervised methods on the Occ3D-nuScenes benchmark (up to a $34\%$ relative improvement).
    Importantly, these gains are consistently observed across multiple plug-and-play occupancy network architectures.
\end{itemize}

\section{Related Work}
\subsection{3D Occupancy Estimation}
The 3D semantic occupancy estimation task has garnered significant attention in autonomous driving research.
Broadly, vision-based approaches can be categorized into fully supervised and shelf-supervised methods.

\textbf{Fully Supervised Methods} leverage dense 3D ground truth annotations, typically derived from LiDAR sensors with additional manual labeling.
Most approaches adapt architectures originally developed for object detection, lifting multi-view camera features into a 3D voxel grid, refining them through 3D convolutions or deformable attention, and applying losses directly in voxel space~\cite{huang2023tri, huang2022bevdet4d, tong2023scene, cao2022monoscene, li2023voxformer}.
Recent advancements have improved the model efficiency~\cite{yu2023flashocc, wang2024opus, lu2023octreeocc, liu2024fully, shi2025occupancy, tang2024sparseocc}, optimized training procedures~\cite{pan2023renderocc, boeder2025occflownet, gan2023simple, hayler2024s4c, sun2024gsrender, huang2024gaussianformer}, or improved the overall occupancy estimation performance through refined architectural designs and specific modeling strategies~\cite{li2023fb, zhang2023occformer, jiang2023symphonize, tan2024geocc, Zhao_2024_CVPR, ma2024cotr, ma2024cam4docc, chen2025alocc, ye2024cvtocc, liao2025stcocc, Chen_2025_ICCV}.
A growing research direction explores the integration of 3D occupancy with feature spaces of foundation models. 
Some methods align predicted voxel features with vision-language representations to enable open-vocabulary occupancy estimation~\cite{vobecky2024pop, yu2024language, tan2023ovo, zheng2025veon, boeder2025langocc, li2025ago}, while~\cite{sirko2024occfeat, jiang2024gausstr} distill DINO features to obtain strong semantic priors.

\textbf{Shelf-Supervised Methods}, on the other hand, aim to mitigate the reliance on expensive 3D labels by utilizing 2D annotations or self-supervision.
SelfOcc~\cite{huang2023selfocc} and OccNeRF~\cite{zhang2023occnerf} employ volume rendering techniques to project estimated 3D occupancy into 2D image space, where photometric and semantic losses from pretrained models~\cite{zhang2023simple, ren2024grounded} provide supervision.
GaussianOcc~\cite{gan2024gaussianocc} and GaussTR~\cite{jiang2024gausstr} instead use 3D Gaussian Splatting for rendering supervision.
GaussianFlowOcc~\cite{Boeder_2025_ICCV} further models scene dynamics to mitigate temporal inconsistencies during training.
While these shelf-supervised methods remove the need for 3D ground truth, they struggle with precise geometric understanding due to the inherent ambiguity of 2D signals and often face challenges in dynamic scenes. 
GS-Occ3D~\cite{Ye_2025_ICCV} recently demonstrated direct 3D Gaussian optimization for occupancy reconstruction, yet without incorporating semantic labels.
LeAP~\cite{gebraad2025leap} proposes a voxel pseudo-label pipeline for occupancy estimation using LiDAR scans as geometric priors.
The concurrent work and preprint EasyOcc~\cite{hayes2025easyocc} instead lifts 2D semantic segmentation masks into 3D using (only) monocular depth estimates, but does not achieve a notable performance improvement over previous work.
It is important to note that some methods, such as AGO~\cite{li2025ago}, AutoOcc~\cite{zhou2025autoocc}, and VEON~\cite{zheng2025veon}, while not using semantic 3D annotations for training, still utilize LiDAR data for geometric supervision, which sets them apart from the purely vision-based, shelf-supervised paradigm we focus on.
Although these methods are not directly comparable to ours, we include a quantitative comparison in the supplement for completeness.
Furthermore, some research is moving towards 4D occupancy and motion prediction, and even world models, incorporating occupancy for future state prediction~\cite{ma2024cam4docc, feng2025gaussian, wang2024occsora, zheng2024occworld}.
Finally, there is considerable amount of work on 4D scene reconstruction for driving scenes, enabling tasks like novel view synthesis, scene editing and simulation~\cite{song2025coda, yan2024street, lu2024drivingrecon, wu20244d, huang2024textit}.

\subsection{Vision Foundation Models}
The rapid progress of deep learning has led to the emergence of powerful Vision Foundation Models (VFMs), which have transformed a wide range of computer vision tasks.
Models such as DINO~\cite{caron2021emerging, oquab2023dinov2} and CLIP~\cite{radford2021learning} are trained on large-scale image corpora to learn general-purpose visual representations.
In parallel, Vision-Language Models (VLMs) combine visual perception with natural language understanding.
Grounding DINO~\cite{liu2024grounding} and Detic~\cite{zhou2022detecting} are prominent examples of open-vocabulary object detectors capable of localizing objects based on arbitrary text prompts.
GroundedSAM~\cite{ren2024grounded} extends this concept by integrating Grounding DINO with the Segment Anything Model (SAM)~\cite{kirillov2023segany}, enabling open-vocabulary segmentation by generating pixel-accurate masks for any textual query without additional training.
Text4Seg~\cite{lan2024text4seg} further leverages multi-modal language models to enhance open-vocabulary segmentation capabilities.

Recently Geometry Foundation Models emerged, which specialize in inferring 3D geometric information from 2D images.
Models such as DUST3R~\cite{wang2024dust3r}, MAST3R~\cite{leroy2024grounding}, VGGT~\cite{wang2025vggt}, and MapAnything~\cite{keetha2025mapanything} are trained on large-scale datasets with 3D supervision (e.g., structure-from-motion data, synthetic scenes) and can estimate depth, camera poses, and dense 3D point clouds.
MapAnything~\cite{keetha2025mapanything}, in particular, is designed for large-scale scene reconstruction and can leverage known camera parameters, making it highly suitable for autonomous driving applications.
These models provide a powerful source of metrically consistent 3D geometry from standard camera inputs, which is foundational to our approach for generating 3D pseudo-labels.

\section{Methodology}\label{sec:method}
Our proposed \paperName{} framework generates high-fidelity, metrically-scaled 3D semantic voxel pseudo-labels from multi-view image sequences.
These pseudo-labels act as a plug-and-play source of supervision for any occupancy estimation network that leverages 3D labels.
The framework is specifically designed to overcome the limitations of directly applying 3D geometry foundation models (FMs) to dynamic scenes.

\begin{figure*}
    \centering
    \includegraphics[page=1, trim=0cm 1.9cm 3.1cm 0cm, clip, width=\textwidth]{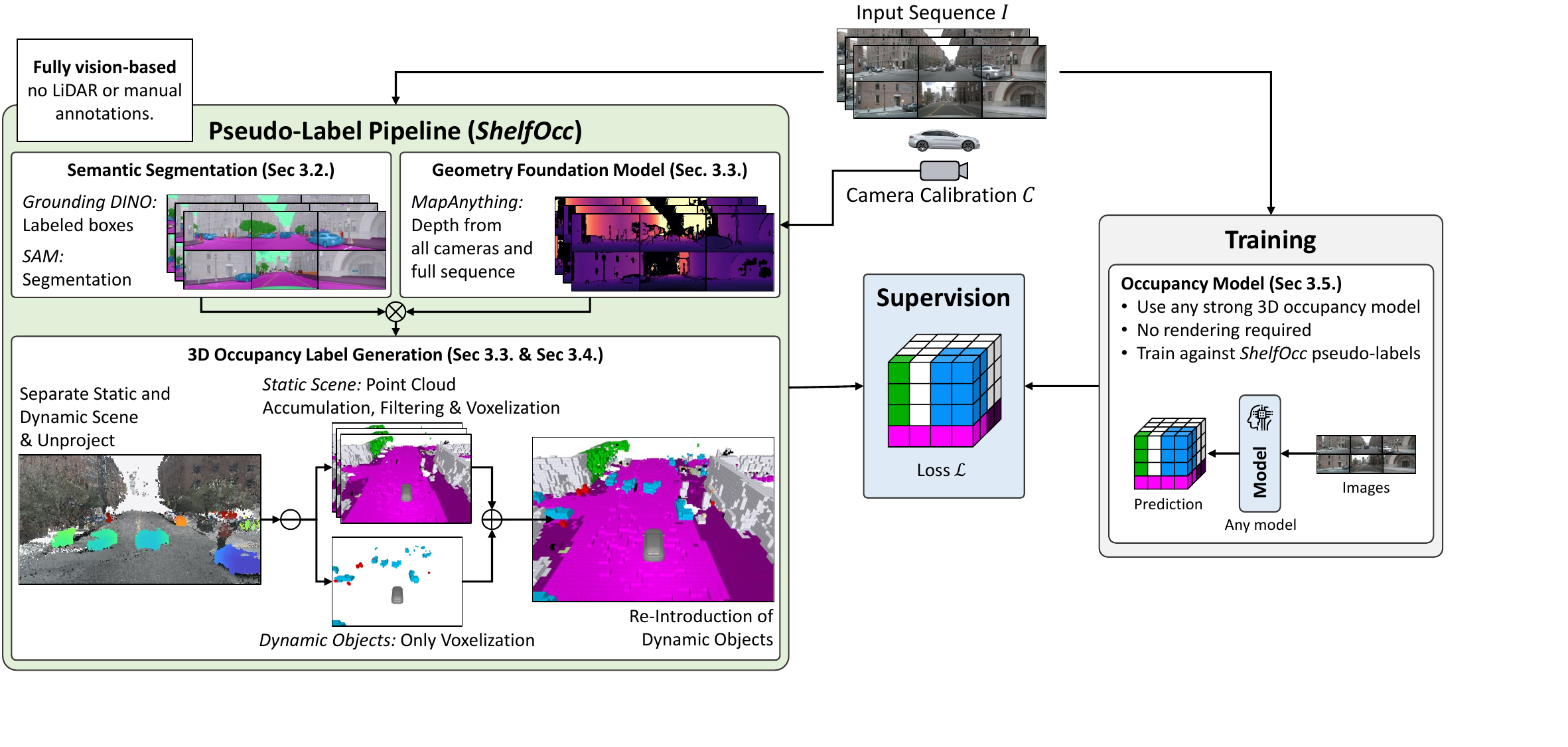}
    \caption{
    \textbf{Overview of the \textit{\paperName{}} framework.}
    We leverage a 3D geometry foundation model (MapAnything~\cite{keetha2025mapanything}) and a 2D semantic foundation model (GroundedSAM~\cite{ren2024grounded}) to construct precise 3D semantic voxel pseudo-labels.
    The pipeline processes image sequences, segregating static and dynamic scene elements, filtering and aggregating static elements and carefully reintroducing dynamic objects to mitigate artifacts.
    These generated 3D pseudo-labels serve as a plug-and-play supervision for any 3D occupancy network.}
    \label{fig:pipeline}
    \vspace{-2mm}
\end{figure*}

The overall framework is illustrated in \cref{fig:pipeline}.
We leverage the 3D geometry model MapAnything~\cite{keetha2025mapanything} and the 2D semantic segmentation model GroundedSAM~\cite{ren2024grounded} to construct semantics-aware 3D pseudo-labels.
The framework comprises six key stages: 2D semantic segmentation mask prediction, initial 3D geometry estimation and filtering, static scene accumulation, dynamic object reintroduction, voxelization with visibility mask generation, and training an occupancy network using the generated labels.
We describe each stage in detail in the following subsections.

\subsection{Notation}
We denote the dataset as a collection of driving sequences 
$\mathcal{S} = \{S_1, S_2, \dots, S_N\}$, 
where each sequence $S_k$ represents a short temporal segment composed of multiple consecutive frames. 
Formally, a sequence $S_k$ contains 
$\mathcal{F}_k = \{f_{1}, f_{2}, \dots, f_{T_k}\}$, 
where $T_k$ is the total number of frames in sequence $S_k$.
Pseudo-label generation is performed independently for each sequence.
At a given time step $t$, the frame $f_t$ consists of a set of multi-view images
$\mathcal{I}_t = \{I_{1,t}, I_{2,t}, \dots, I_{C,t}\}$,
where $C$ denotes the total number of cameras. 
Each image $I_{i,t}$ is associated with its intrinsic calibration matrix $\mathbf{K}_{i,t}$ and extrinsic camera pose $\mathbf{T}_{i,t}$, which together define the projection from 3D world coordinates to the 2D image plane and vice versa.

\subsection{2D Pseudo-Semantic Masks}\label{sec:semseg}
The initial step in the \paperName{} pipeline involves generating dense 2D pseudo-semantic segmentation masks for all input images $I_i$ within the sequence.
We employ GroundedSAM~\cite{ren2024grounded}, which combines an open-vocabulary object detector and a highly capable segmentation model.
The process unfolds in two distinct stages.
Firstly, each image is fed into Grounding DINO~\cite{liu2024grounding}, an open-vocabulary object detection method, along with a set of text prompts corresponding to the target semantic classes.
The predicted bounding boxes then serve as input prompts for SAM (Segment Anything Model)~\cite{kirillov2023segany}, which then generates precise 2D masks, inheriting the label from its corresponding predicted box.

\paragraph{Mitigating False Negatives / Positives via sky grounding.} We observed that querying Grounding DINO for all target classes simultaneously often results in too few detections, causing many objects in the image to be missed.
However, querying the model for each class individually can lead to a bias towards detecting objects even when they are not present, resulting in high-confidence but incorrect boxes and frequent class confusions.
To address this issue, we input each class prompt individually into Grounding DINO, together with a generic background label (e.g., ‘sky’).
This strategy greatly reduces false positives by providing Grounding DINO an alternative high-confidence prediction when the query object is absent.
Any boxes predicted with the background label are discarded before the second stage with SAM.
The full vocabulary used for generating the semantic segmentation masks can be found in the supplementary material.
These dense 2D masks are crucial for assigning semantic information to the 3D points and for dynamically identifying objects within the scene, a critical step detailed in \cref{sec:method_geometry}.

\subsection{3D Geometry Estimation and Static/Dynamic Separation}\label{sec:method_geometry}
We process the entire multi-view image sequence (across all cameras $C$ and time steps $T$) through a 3D geometry foundation model.
For this work, we specifically leverage MapAnything~\cite{keetha2025mapanything}.
MapAnything is particularly suitable because, unlike some other FMs, it can optionally take available camera poses (intrinsics $\mathbf{K}_i$ and extrinsics $\mathbf{T}_i$) as input, which are readily available from autonomous driving datasets like nuScenes~\cite{caesar2020nuscenes}.
This enables MapAnything to directly predict metrically-scaled depth maps $\mathbf{D}_i$ for each input image $I_i$.
The depth maps can be unprojected using the camera poses and intrinsics to create a 3D point cloud representing the scene geometry.
One of the primary challenges in generating accurate 3D pseudo-labels for dynamic driving scenes is the handling of moving objects.
Naively accumulating points from all time steps of the sequence into a single 3D scene would result in moving objects appearing multiple times along their trajectory, polluting the scene representation.
To prevent this, we separate the scene into static and a dynamic components.

\paragraph{Static Scene Construction.} To construct the static part of the scene, which remains consistent across all time steps within a sequence, we first use the 2D pseudo-semantic masks from \cref{sec:semseg} to determine which pixels belong to dynamic objects.
We then unproject only those pixels that are \textit{not} identified as dynamic into 3D points.
This selective unprojection prevents motion artifacts, which are common because FMs like MapAnything are primarily trained on static scenarios.
The 3D coordinate $\mathbf{P}(u,v)$ for a pixel $(u,v)$ in image $\mathbf{I}_i$ are obtained by:
\begin{equation}
    \mathbf{P}(u,v) = \mathbf{T}_i \cdot \left( \mathbf{K}_i^{-1} \cdot \begin{pmatrix} u \\ v \\ 1 \end{pmatrix} \cdot \mathbf{D}_i(u,v) \right).
    \label{eq:unproject}
\end{equation}
Here, $\mathbf{D}_i(u,v)$ is the depth predicted by MapAnything, $\mathbf{K}_i$ represents the camera intrinsics, and $\mathbf{T}_i$ are the camera extrinsics.
All static points, collected from all cameras and time steps within a given sequence, are then aggregated into a single global static point cloud $\mathcal{P}_{\text{static}}$.

\paragraph{Confidence Filtering.}\label{sec:filter} To mitigate noise in the accumulated static point cloud and suppress spurious occupied voxels in the final 3D volume, we apply two confidence filtering strategies.
First, we remove points likely resulting from erroneous depth predictions by retaining only those that are confirmed by multiple frames within the sequence.
Concretely, for each pixel ray, we record how often it traverses a given voxel cell and how often it terminates within that cell based on the unprojected depth points.
Points located in cells that are intersected more frequently than they are confirmed are discarded, as they likely correspond to incorrect depth estimates.
Second, we prune voxels with insufficient point density (fewer than four points in our case), as such sparsely populated regions tend to represent unreliable or noisy predictions.
Together, these filtering steps substantially improve the consistency and reliability of the generated point cloud, preserving only high-confidence geometric information for subsequent voxelization.

\paragraph{Dynamic Scene Construction.} For each time step $t$ in the sequence, we generate a dedicated dynamic point cloud $\mathcal{P}_{\text{dynamic}, t}$.
This is achieved by unprojecting all pixels that were identified as dynamic by the 2D semantic segmentation in all cameras $\mathbf{I}_{i,t}$ at that time step using \cref{eq:unproject}.
This ensures that dynamic objects are captured precisely at their positions in each frame.

\subsection{Voxelization and Visibility Mask Generation}
The final 3D semantic point cloud for any given frame $t$ is then constructed by combining the global static point cloud with the dynamic point cloud specific to that time step.
The final step in the \paperName{} pipeline converts the 3D semantic point clouds $\mathcal{P}_t$ into high-fidelity, metrically-scaled 3D semantic voxel labels, which are ready for direct supervision.

\paragraph{Voxelization.} A target voxel grid is first defined with specific dimensions and resolution (e.g., $[-40m, 40m]$ in X/Y and $[-1m, 5.4m]$ in Z with a $0.4m$ resolution for nuScenes).
For each voxel $\mathbf{v}$ in this grid, we aggregate all 3D points from $\mathcal{P}_t$ that fall within its spatial bounds.
The semantic label of $\mathbf{v}$ is then determined by a majority voting scheme among these collected points.
To mitigate class imbalance effects near the ground where frequent classes (\emph{road} or \emph{terrain}) tend to dominate, we prioritize minority object classes (e.g., \emph{traffic cones}).
If no points fall within a voxel, it is marked as \emph{empty}.
This process yields a dense 3D semantic voxel grid $\mathbf{V}_t$ for each frame in the sequence $S_i$.

\paragraph{Camera Visibility Mask.} To effectively train an occupancy network, it is essential to distinguish between truly empty space and regions that are merely unobserved by the cameras.
For each frame $t$, we generate a camera visibility mask $\mathbf{M}_{\text{vis},t}$.
This is achieved by casting rays from each ground truth camera position through the scene.
For each ray, we identify the first occupied voxel.
All voxels along the ray \textit{before} this first occupied voxel are marked as 'visible free space'.
Conversely, voxels that lie behind occupied voxels or are entirely outside the frustum of any camera are marked as 'unobserved'.
This visibility mask $\mathbf{M}_{\text{vis},t}$ is critical during network training, as it guides the loss computation, ensuring that the model is only penalized for incorrect predictions within regions theoretically observable by the cameras.
This accurate voxelization process ultimately provides high-quality, dense 3D semantic voxel grids $\mathbf{V}_t$ and corresponding visibility masks $\mathbf{M}_{\text{vis},t}$.

\newcommand{\snd}[1]{\textit{#1}}
\begin{table*}[!ht]
	\begin{center}
		\caption{
			\textbf{Occupancy estimation performance on the Occ3D-nuScenes validation set.}
			Best performing per column in \textbf{bold}, second best in \textit{italics}.
			All methods ignore the \textit{others} and \textit{other flat} classes.
            \textit{IoU} represents the geometric performance independent of the semantic label, while \textit{mIoU} is the mean IoU over all classes.
        }
		\label{table:main_complete}
		
		\resizebox{\textwidth}{!}{%
			\addtolength{\tabcolsep}{2pt}
			\begin{tabular}{l|cc|ccccccccccccccc}
				\hline
				\noalign{\smallskip}
				Method & mIoU & IoU & 
                \rotatebox{90}{\textcolor{barrier}{$\blacksquare$} barrier} &
                \rotatebox{90}{\textcolor{bicycle}{$\blacksquare$} bicycle} &
                \rotatebox{90}{\textcolor{bus}{$\blacksquare$} bus} &
                \rotatebox{90}{\textcolor{car}{$\blacksquare$} car} &
                \rotatebox{90}{\textcolor{construction}{$\blacksquare$} cons. vehicle} &
                \rotatebox{90}{\textcolor{motorcycle}{$\blacksquare$} motorcycle} &
                \rotatebox{90}{\textcolor{pedestrian}{$\blacksquare$} pedestrian} &
                \rotatebox{90}{\textcolor{cone}{$\blacksquare$} traffic cone} &
                \rotatebox{90}{\textcolor{trailer}{$\blacksquare$} trailer} &
                \rotatebox{90}{\textcolor{truck}{$\blacksquare$} truck} &
                \rotatebox{90}{\textcolor{driveable}{$\blacksquare$} driv. surf.} &
                \rotatebox{90}{\textcolor{sidewalk}{$\blacksquare$} sidewalk} &
                \rotatebox{90}{\textcolor{terrain}{$\blacksquare$} terrain} &
                \rotatebox{90}{\textcolor{manmade}{$\blacksquare$} manmade} &
                \rotatebox{90}{\textcolor{vegetation}{$\blacksquare$} vegetation}\\
				
				\noalign{\smallskip}
				\hline
				\noalign{\smallskip}
				
				SelfOcc~\cite{huang2023selfocc} & 10.54 & 44.05 & 0.15 & 0.66 & 5.46 & 12.54 & 0.00 & 0.80 & 2.10 & 0.00 & 0.00 & 8.25 & 55.49 & 26.30 & 26.54 & 14.22 & 5.60 \\
				
				OccNeRF~\cite{zhang2023occnerf} & 10.81 & 46.43 & 0.83 & 0.82 & 5.13 & 12.49 & 3.50 & 0.23 & 3.10 & 1.84 & 0.52 & 3.90 & 52.62 & 20.81 & 24.75 & 18.45 & 13.19 \\
				
				GaussianOcc~\cite{gan2024gaussianocc} & 11.26 & 42.91 & 1.79 & 5.82 & 14.58 & 13.55 & 1.30 & 2.82 & 7.95 & 9.76 & 0.56 & 9.61 & 44.59 & 20.10 & 17.58 & 8.61 & 10.29 \\
				
				GaussTR~\cite{jiang2024gausstr} & 13.26 & 44.54 & 2.09 & 5.22 & 14.07 & 20.43 & \snd{5.70} & 7.08 & 5.12 & 3.93 & 0.92 & 13.36 & 39.44 & 15.68 & 22.89 & 21.17 & \snd{21.87} \\
				
				EasyOcc~\cite{hayes2025easyocc} & 15.96 & 38.86 & 1.85 & 8.18 & 16.66 & \snd{22.12} & 1.01 & 7.74 & \textbf{14.74} & \snd{12.84} & 0.98 & 13.76 & 55.91 & 27.96 & 22.73 & 15.77 & 17.20 \\
				
				GaussianFlowOcc~\cite{Boeder_2025_ICCV} & 17.08 & 46.91 & 6.75 & \snd{9.68} & 18.98 & 17.15 & 4.19 & \snd{11.78} & 9.27 & 10.30 & \snd{1.83} & 12.33 & 61.03 & 31.17 & 34.78 & 14.66 & 12.40 \\
				
				\noalign{\smallskip}
				\hline
				\noalign{\smallskip}

				Ours: \paperName{} + COTR~\cite{ma2024cotr} & 18.65 & \snd{53.71} & 9.10 & 6.20 & \snd{22.92} & 22.08 & 1.66 & 5.94 & 9.92 & 8.55 & 0.0 & \snd{15.32} & 67.93 & 31.13 & 38.76 & \snd{23.11} & 17.15 \\
				
				Ours: \paperName{} + CVT-Occ~\cite{ye2024cvtocc} & \snd{19.21} & 52.72 & \snd{11.53} & 6.38 & 20.39 & 21.92 & 4.20 & 10.18 & 9.02 & 10.67 & 0.89 & 13.08 & \textbf{68.42} & \snd{31.23} & \snd{41.42} & 22.74 & 16.15 \\
								
				Ours: \paperName{} + STCOcc~\cite{liao2025stcocc} & \textbf{22.87} & \textbf{56.14} & \textbf{13.98} & \textbf{11.36} & \textbf{25.27} & \textbf{25.80} & \textbf{7.25} & \textbf{16.61} & \snd{12.91} & \textbf{13.42} & \textbf{5.37} & \textbf{17.15} & \snd{68.01} & \textbf{34.66} & \textbf{42.73} & \textbf{25.63} & \textbf{22.89} \\
				
				\noalign{\smallskip}
				\hline
			\end{tabular}
			\addtolength{\tabcolsep}{2pt}
		}
	\end{center}
	\vspace{-5mm}
\end{table*}

\subsection{Training with \paperName{} Labels}
The central advantage of \paperName{} is its modular design, enabling seamless integration across different occupancy prediction algorithms.
Any existing occupancy network architecture that uses 3D voxel labels for supervision can be directly trained using \paperName{} labels.
This circumvents the need for complex differentiable rendering pipelines that are typically employed in 2D-supervised methods.

An occupancy network takes multi-view images as input and predicts a dense 3D semantic voxel grid $\mathbf{\hat{V}}_t$, representing semantic probabilities for each voxel.
This output is directly compared with the generated \paperName{} labels $\mathbf{V}_t$ using loss functions like the cross-entropy loss 
\begin{equation}
    \mathcal{L} = \sum_{t} \sum_{\mathbf{v} \in \mathbf{V}_t} \mathbf{M}_{\text{vis},t}(\mathbf{v}) \cdot \mathcal{L}_{\text{CE}}(\mathbf{\hat{V}}_t(\mathbf{v}), \mathbf{V}_t(\mathbf{v})),
\end{equation}
however the exact loss functions used depends on the model of choice.
The camera visibility mask $\mathbf{M}_{\text{vis},t}(\mathbf{v})$ is usually applied as a weighting factor in the loss computation.
This ensures that the loss is computed only for voxels within observable regions of the scene.

This direct 3D supervision offers two key advantages: 
First, it significantly enhances 3D geometric understanding by allowing the network to learn directly from explicit 3D targets, effectively mitigating issues such as depth bleeding and promoting a more complete and accurate geometric representation.
Second, it simplifies the training pipeline by eliminating the need for complex differentiable rendering mechanisms, thereby reducing memory consumption and computational overhead.

\section{Experiments}
\subsection{Dataset}
We conduct our experiments on the Occ3D-nuScenes benchmark~\cite{tian2023occ3d}, which builds upon the widely used nuScenes dataset~\cite{caesar2020nuscenes, fong2022panoptic} and provides 3D semantic occupancy ground truth for all scenes in the dataset (that we only use during validation). 
Following standard protocol, we evaluate model performance using the Intersection-over-Union (IoU) across a predefined set of semantic classes, with the mIoU being the mean over all classes. 
In addition, we report the geometric IoU, which measures the voxel-wise accuracy of occupied versus free space regardless of the underlying semantic category, providing a measure of geometric fidelity.
To further assess consistency along the depth axis, we also report the RayIoU metric~\cite{liu2024fully}, a ray-based evaluation that mitigates inconsistencies in traditional voxel-level IoU by comparing occupancy predictions along camera rays. 

\subsection{Experimental Setup}
We focus on the shelf-supervised setting, where models are trained using only multi-view camera inputs without access to LiDAR data or any 3D annotations from the target dataset. 
To assess the quality and generality of our generated 3D pseudo-labels, we train several state-of-the-art occupancy networks originally designed for fully supervised training, namely COTR~\cite{ma2024cotr}, CVT-Occ~\cite{ye2024cvtocc}, and STCOcc~\cite{liao2025stcocc}, directly on our generated pseudo-labels. 
This enables direct comparison to previous shelf-supervised methods while leveraging the advantages of established 3D architectures.
All models are trained with input images at a resolution of $256 \times 704$, using a ResNet-50~\cite{he2016deep} backbone pretrained on ImageNet~\cite{imagenet}. 
Source code will be made available after publication.

\subsection{Main Results}
\begin{figure*}[h]
	\centering
	\makeatletter
	\@wholewidth0.5pt
	\makeatother
	\begin{tabular}{ccccc}
		& \textbf{Input Images} & \textbf{Occ3D GT} & \textbf{\paperName{} Labels} & \textbf{\paperName{} Prediction}\\  
		
		\raisebox{0.15\height}{\rotatebox{90}{\textbf{Scene 0103}}} & 
		\frame{\includegraphics[width=.34\textwidth]{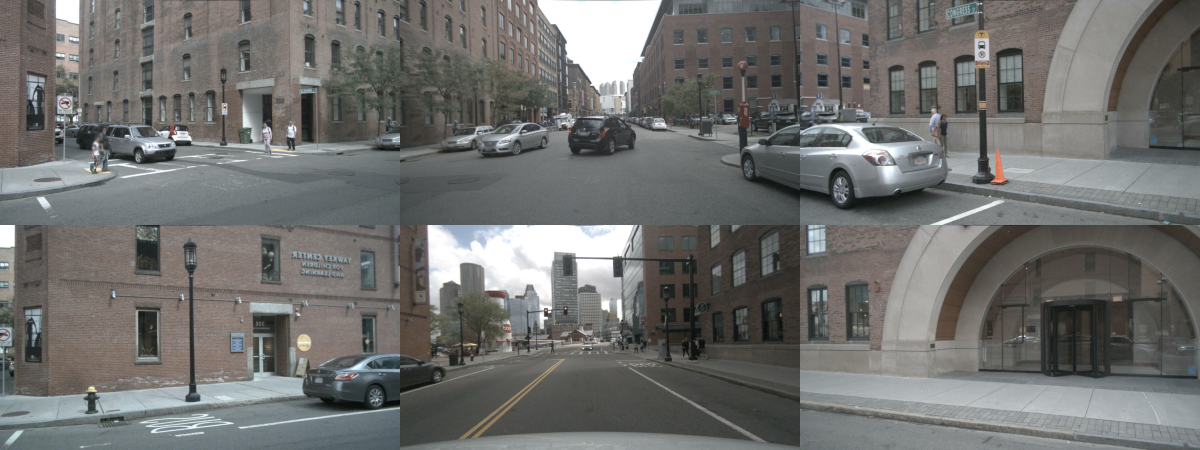}} & 
		\frame{\includegraphics[trim=0cm 0cm 0cm 1cm, clip, width=.175\textwidth]{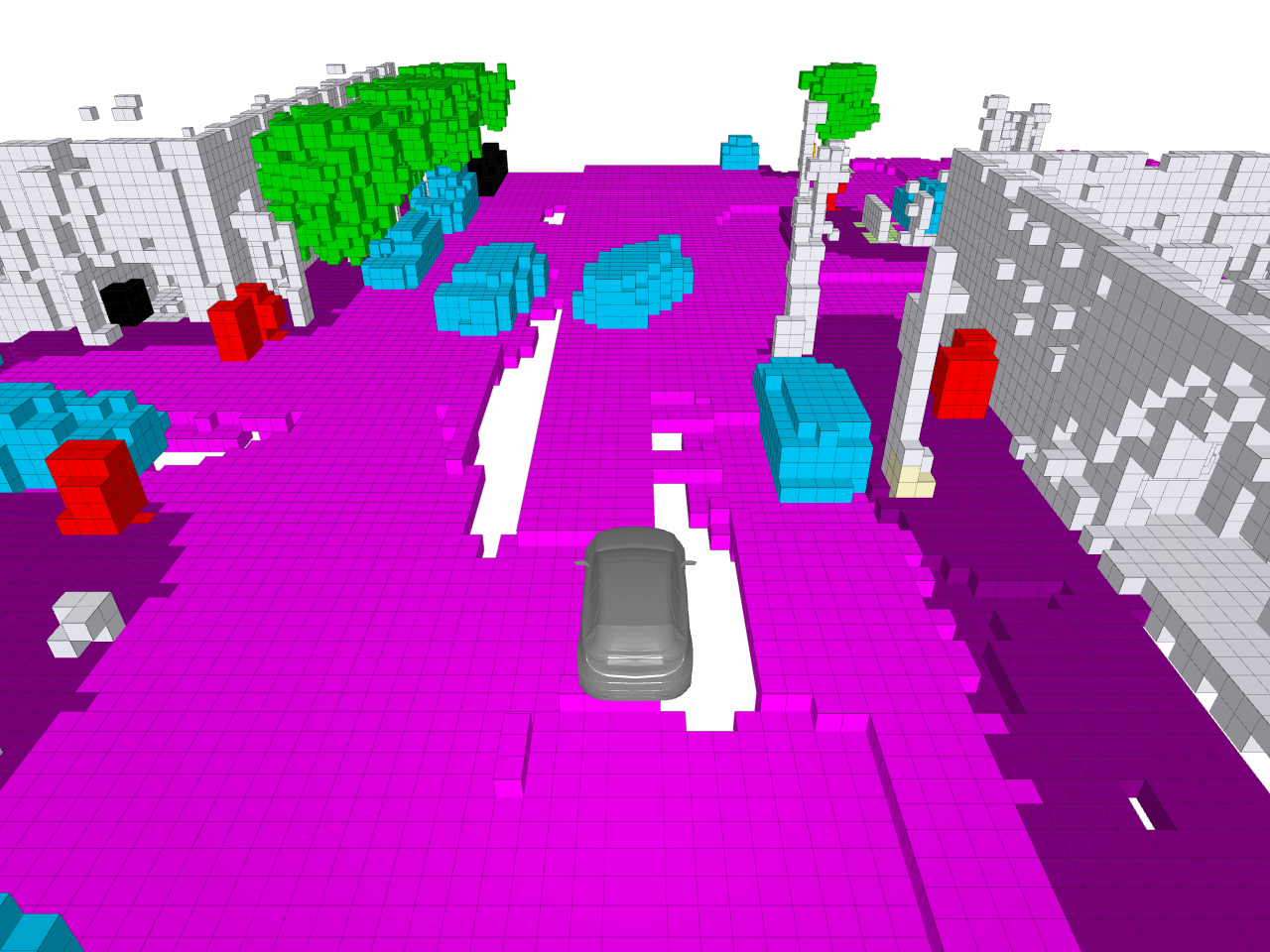}} & 
		\frame{\includegraphics[trim=0cm 0cm 0cm 1cm, clip, width=.175\textwidth]{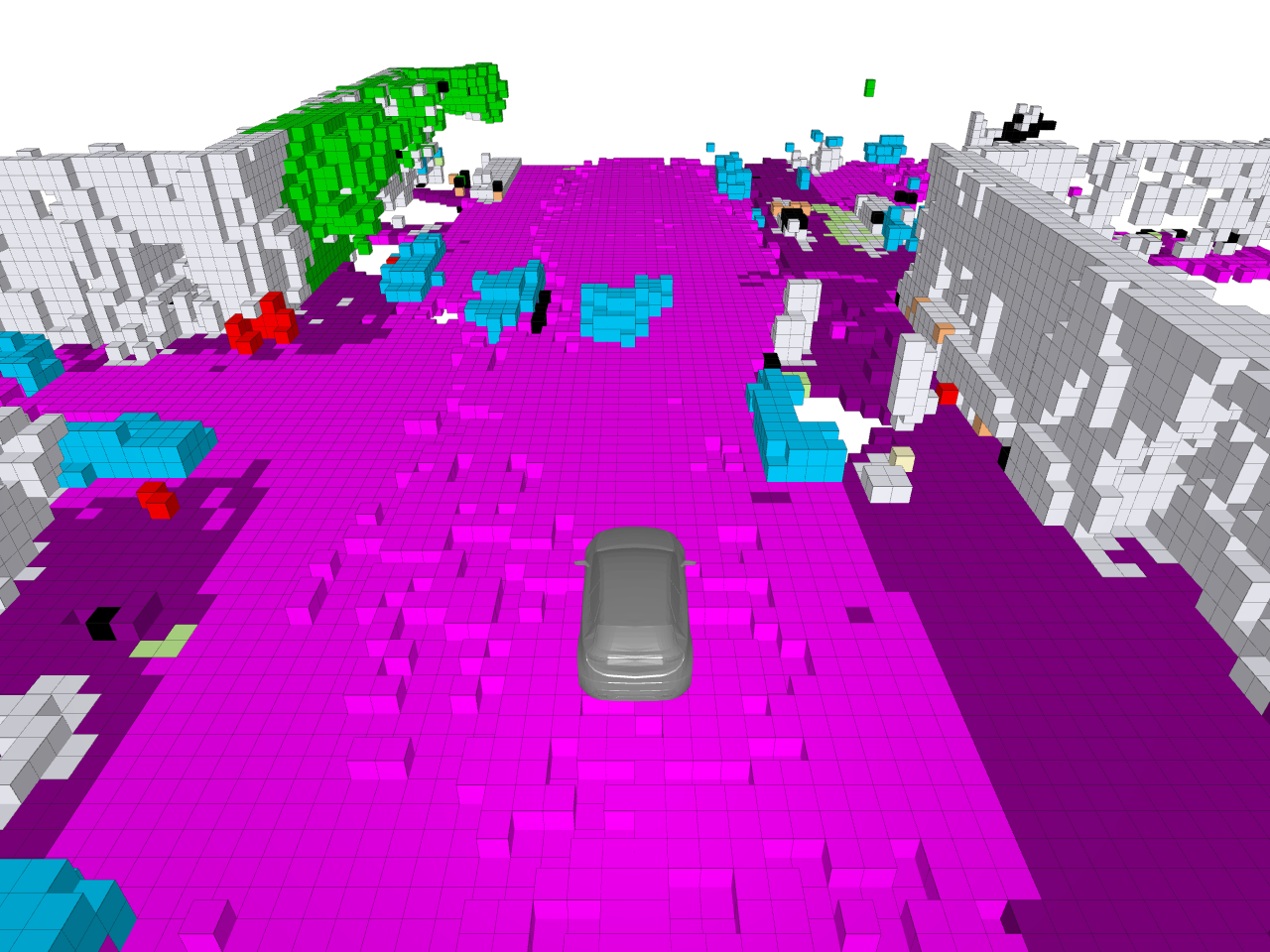}} & 
		\frame{\includegraphics[trim=0cm 0cm 0cm 1cm, clip, width=.175\textwidth]{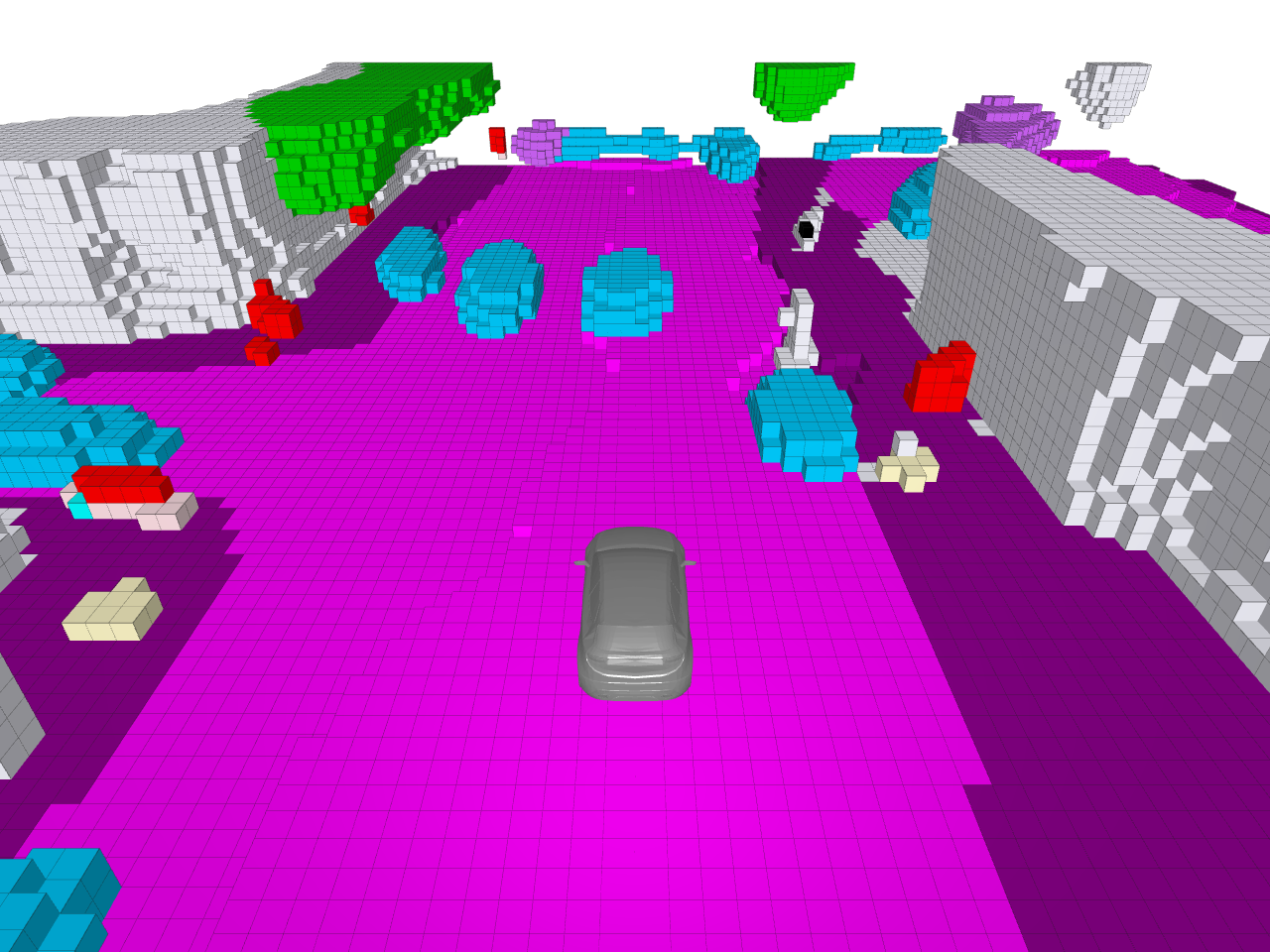}} \\

		\raisebox{0.15\height}{\rotatebox{90}{\textbf{Scene 0521}}} & 
		\frame{\includegraphics[width=.34\textwidth]{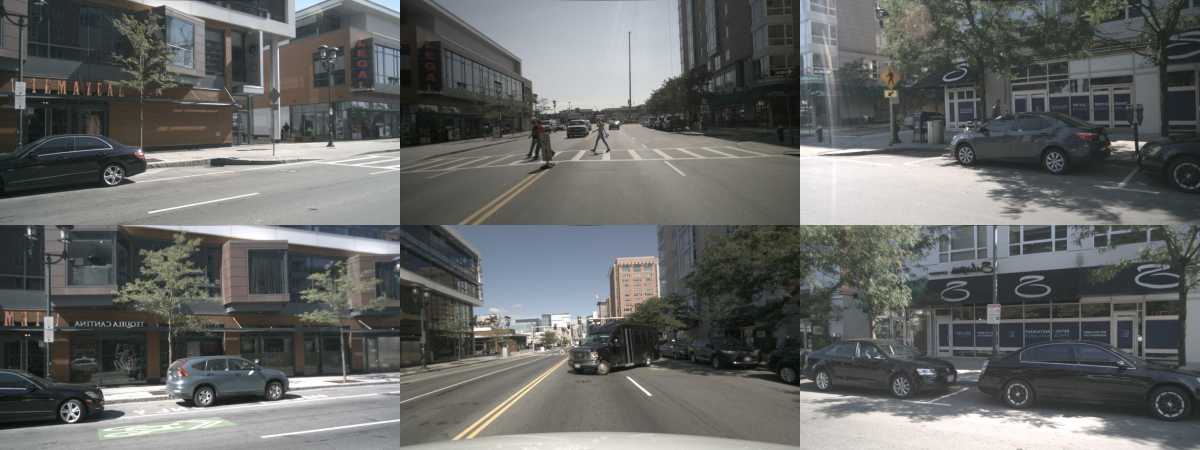}} & 
		\frame{\includegraphics[trim=0cm 0cm 0cm 1cm, clip, width=.175\textwidth]{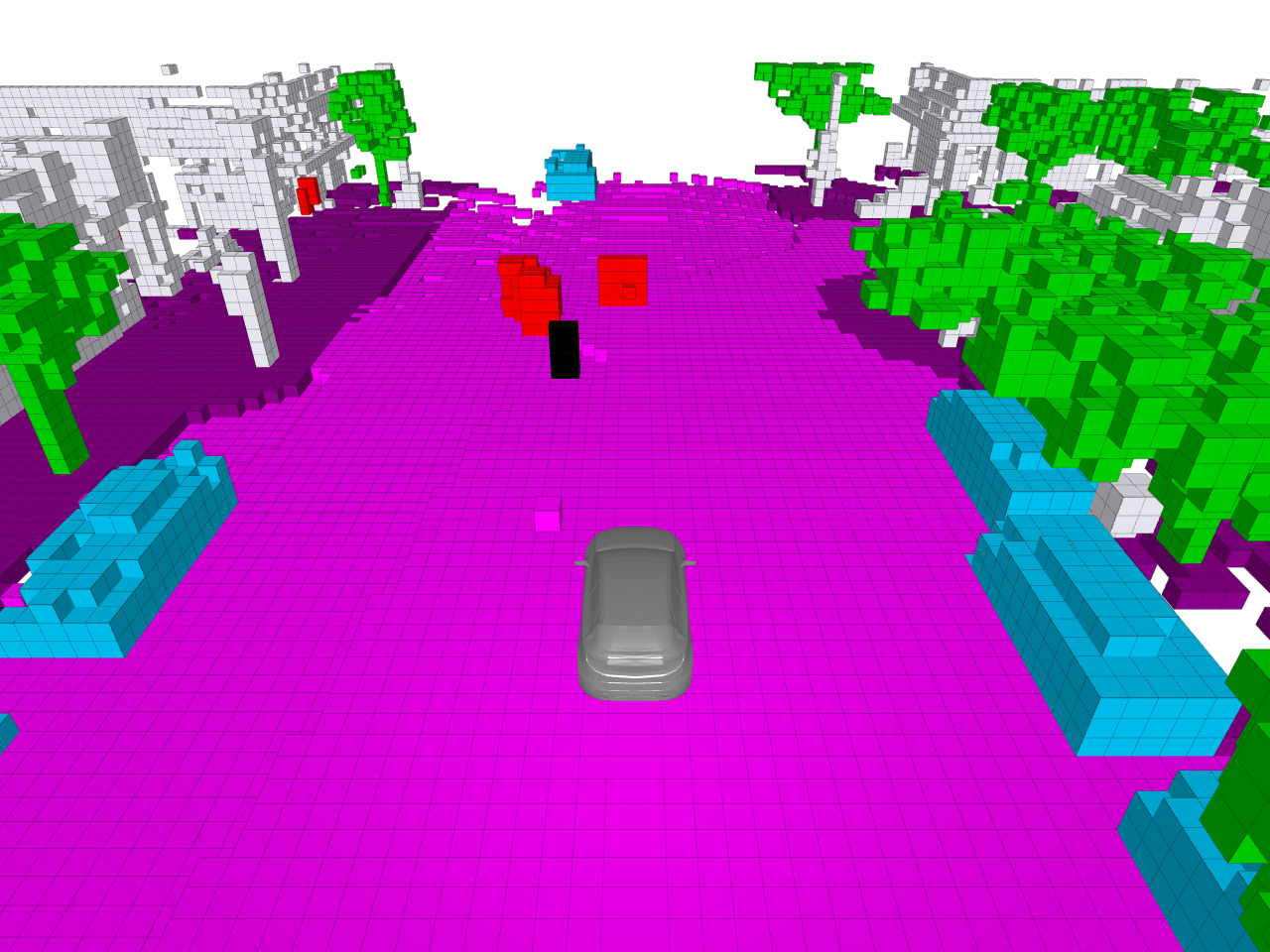}} & 
		\frame{\includegraphics[trim=0cm 0cm 0cm 1cm, clip, width=.175\textwidth]{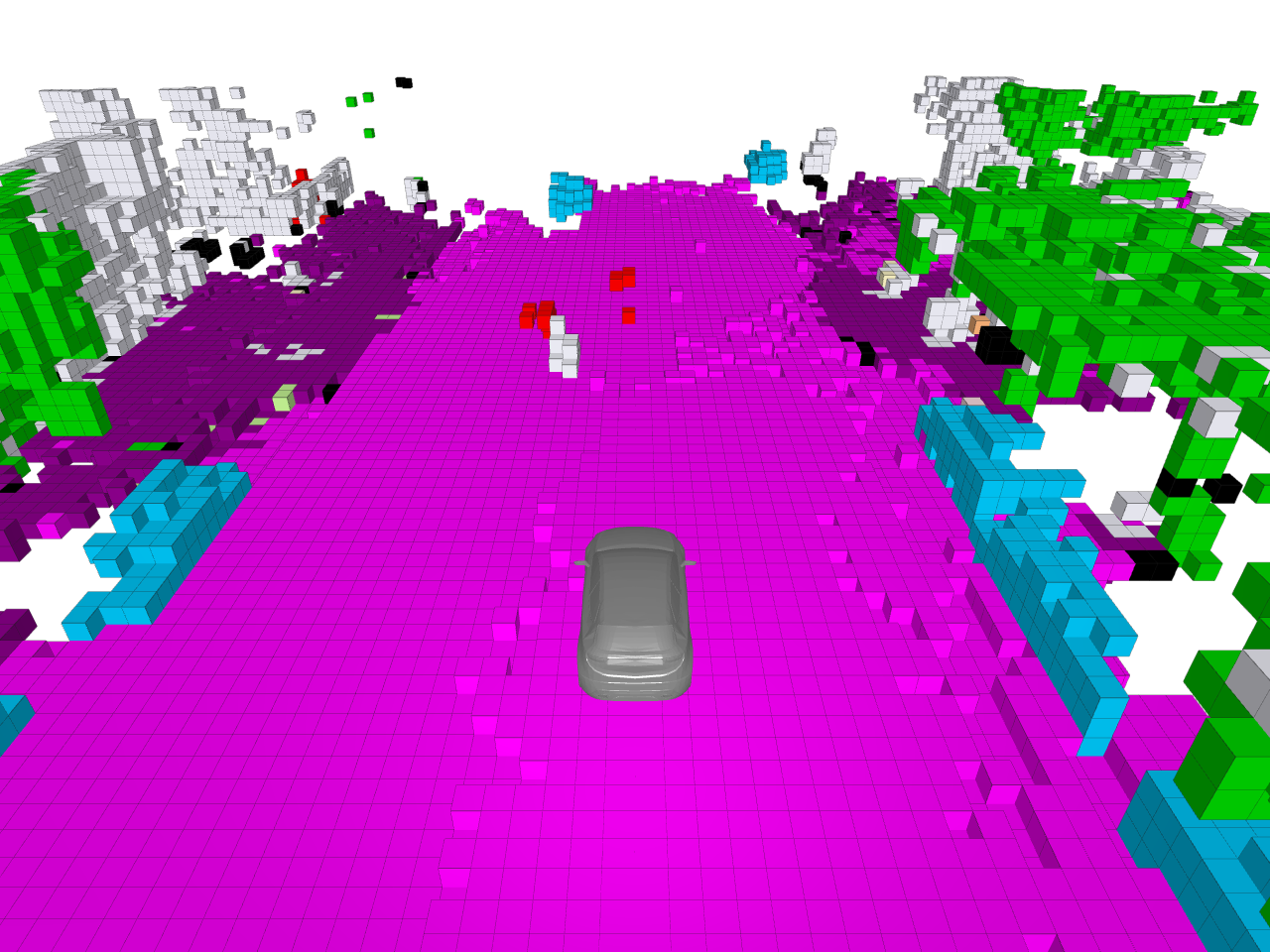}} & 
		\frame{\includegraphics[trim=0cm 0cm 0cm 1cm, clip, width=.175\textwidth]{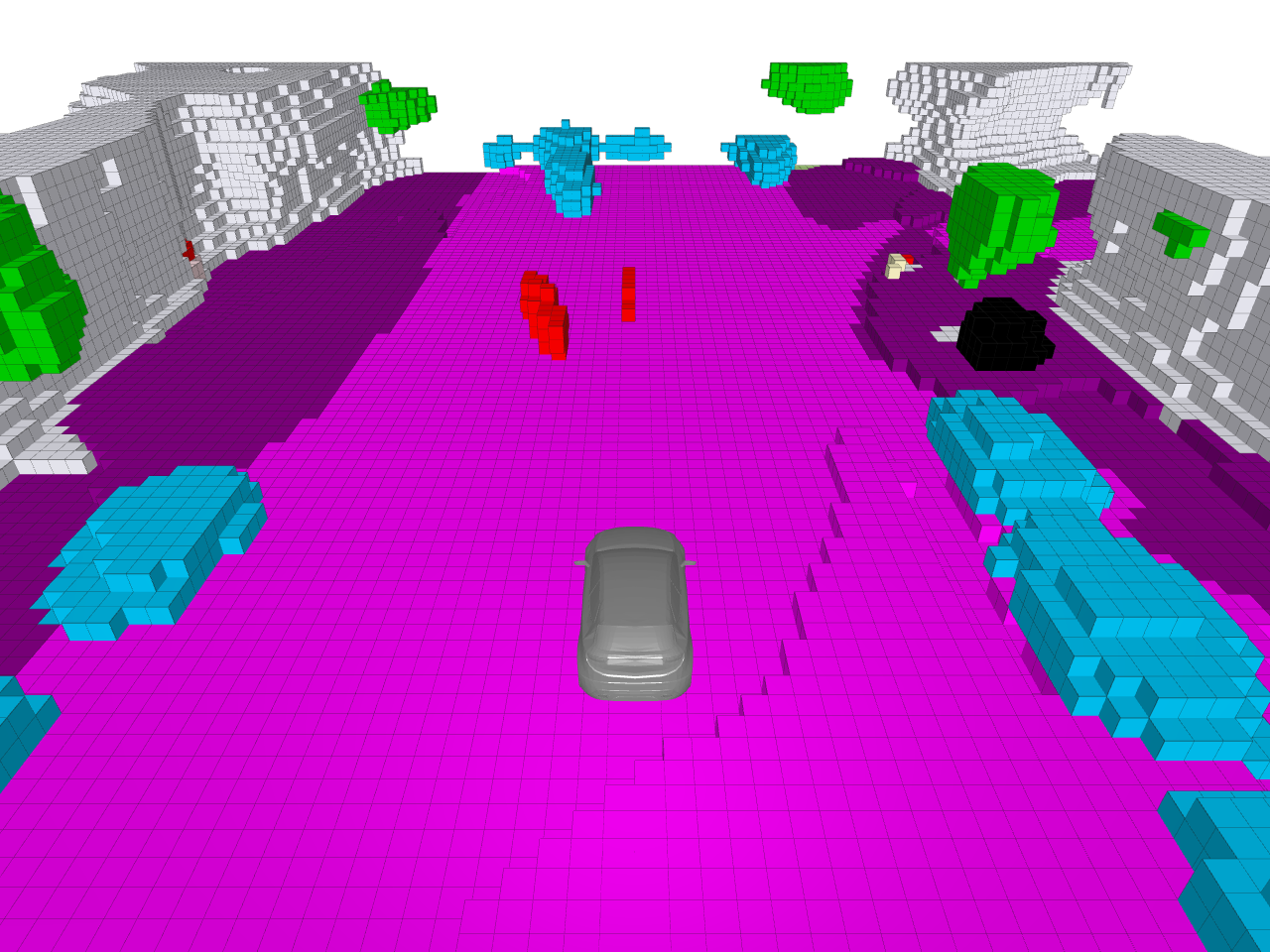}} \\
		
		\raisebox{0.15\height}{\rotatebox{90}{\textbf{Scene 0636}}} & 
		\frame{\includegraphics[width=.34\textwidth]{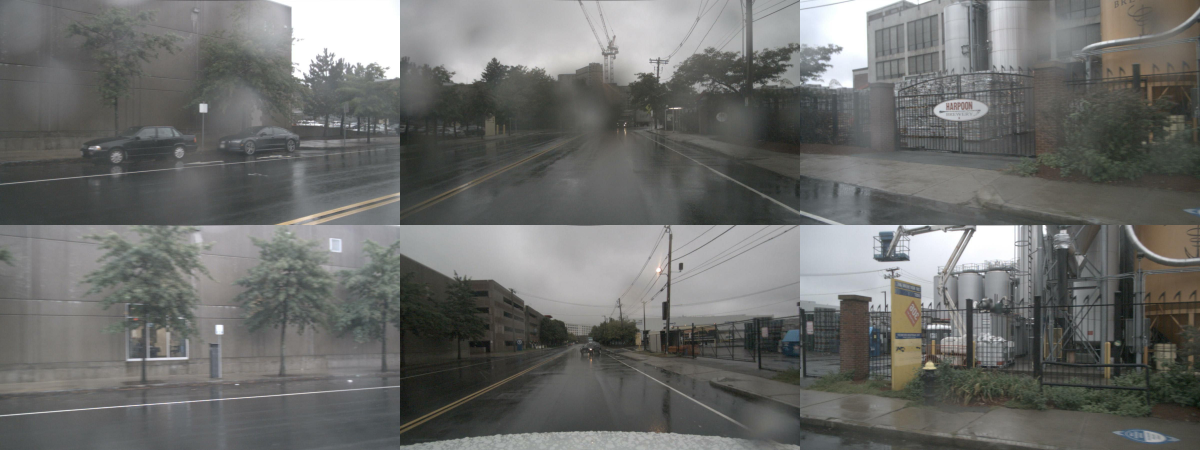}} & 
		\frame{\includegraphics[trim=0cm 0cm 0cm 1cm, clip, width=.175\textwidth]{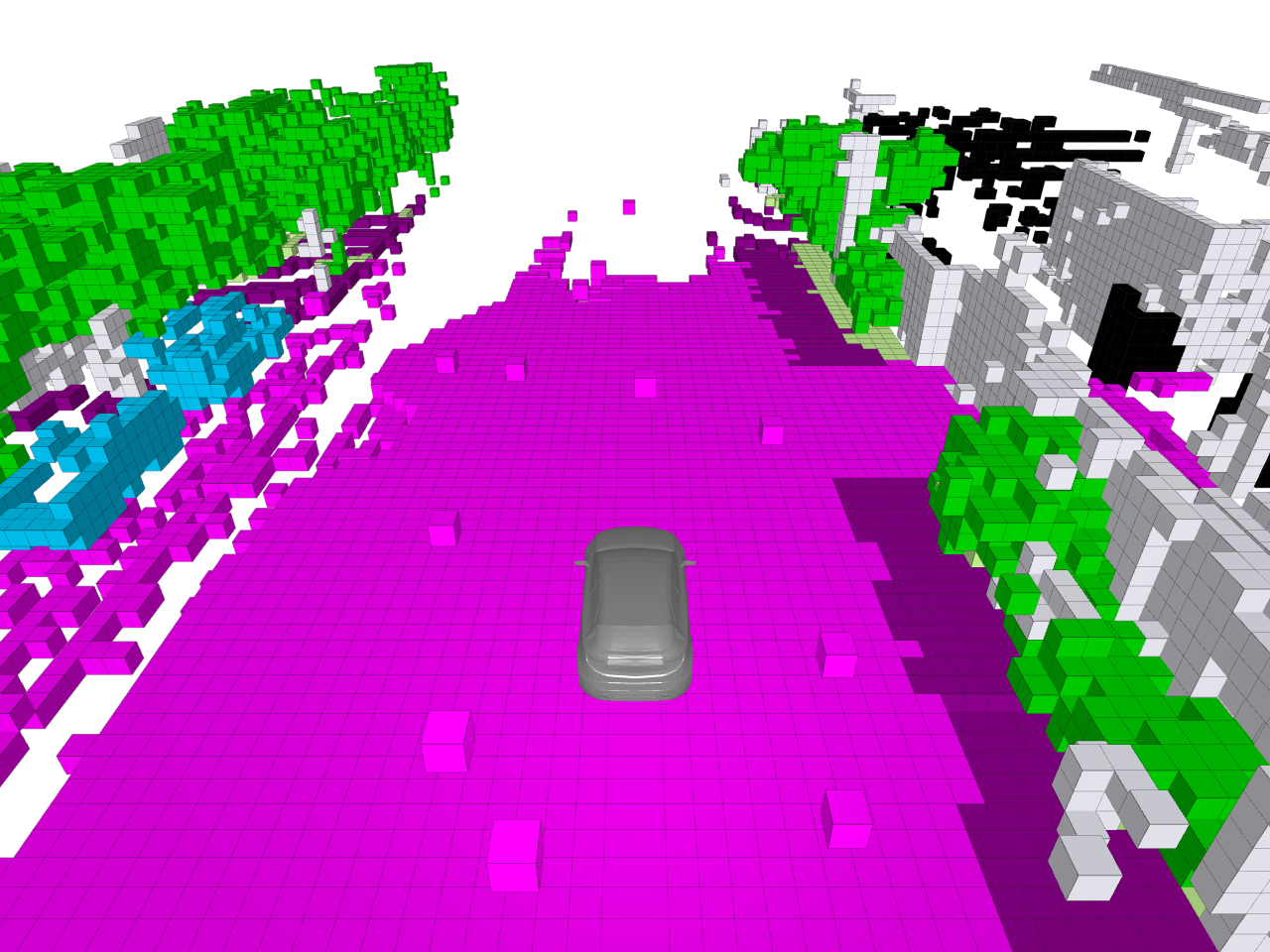}} & 
		\frame{\includegraphics[trim=0cm 0cm 0cm 1cm, clip, width=.175\textwidth]{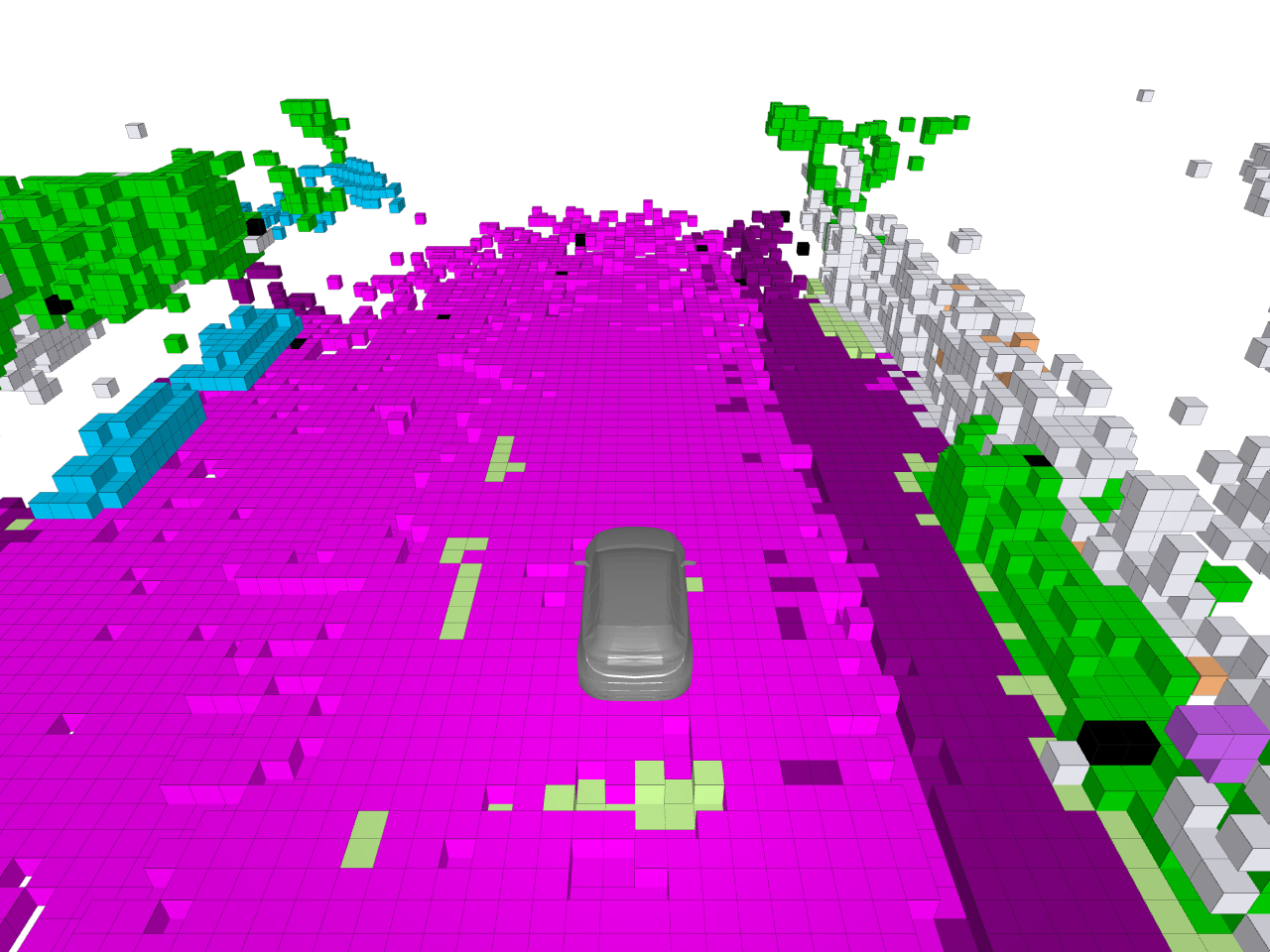}} & 
		\frame{\includegraphics[trim=0cm 0cm 0cm 1cm, clip, width=.175\textwidth]{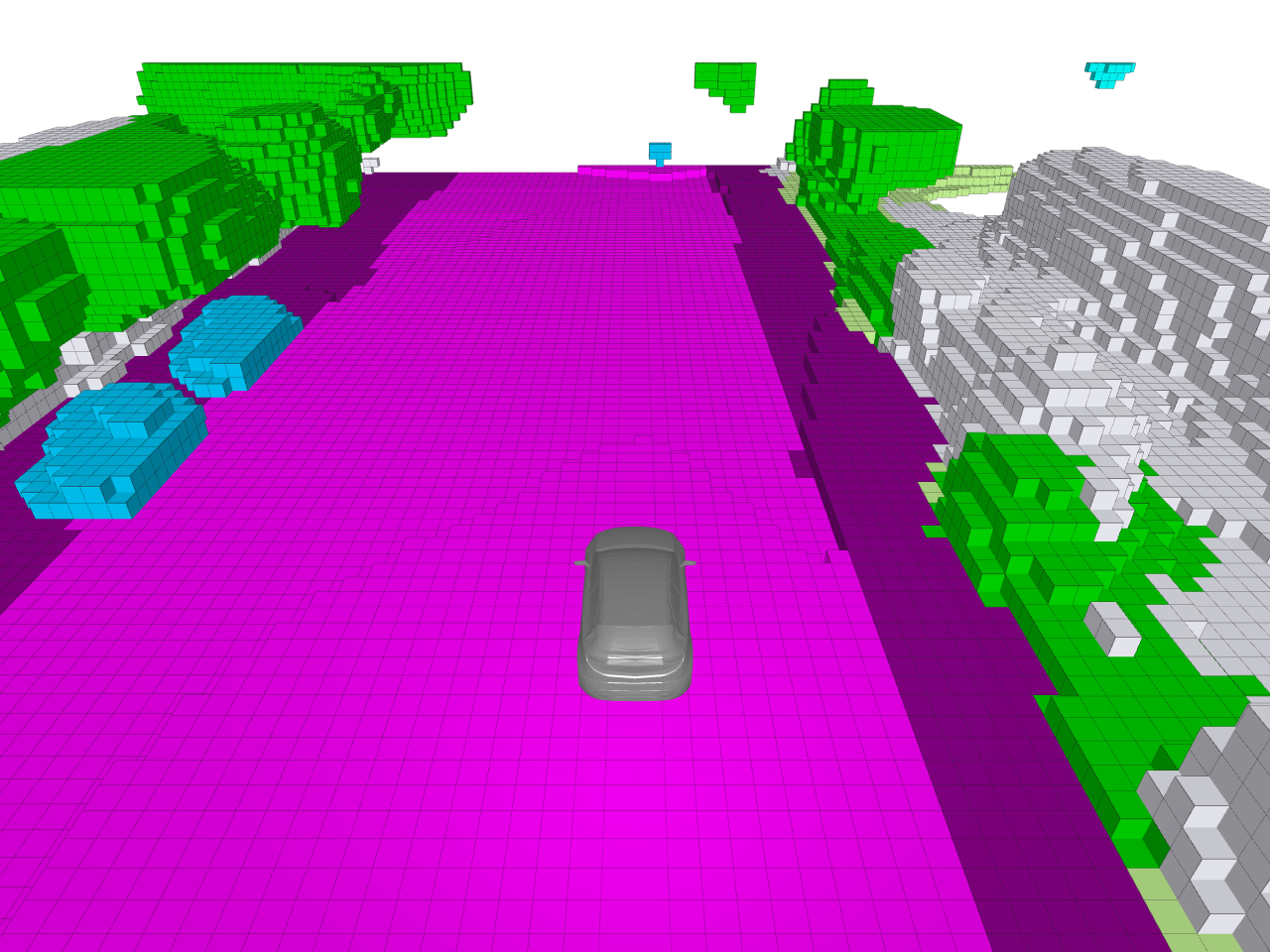}} \\

		\raisebox{0.15\height}{\rotatebox{90}{\textbf{Scene 0916}}} & 
		\frame{\includegraphics[width=.34\textwidth]{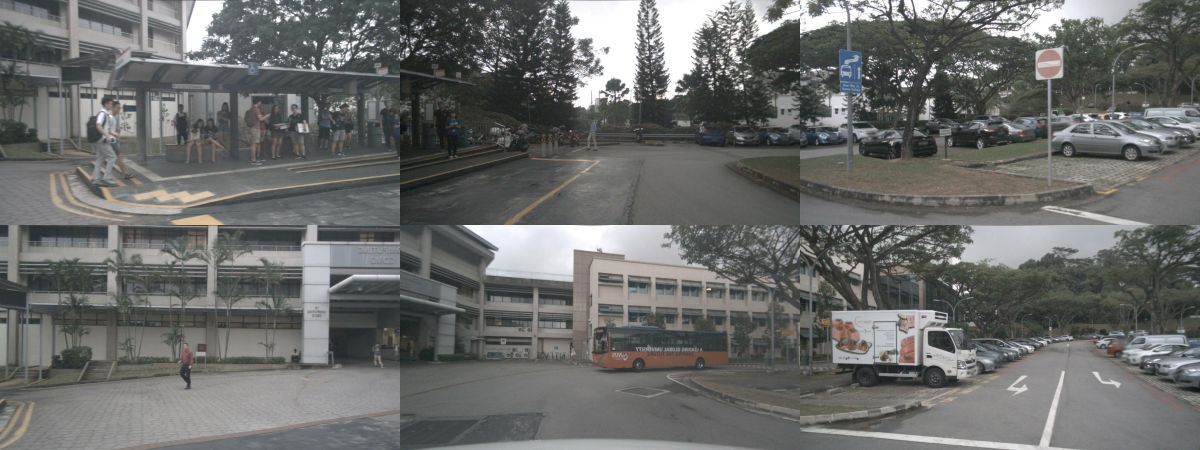}} & 
		\frame{\includegraphics[trim=0cm 0cm 0cm 1cm, clip, width=.175\textwidth]{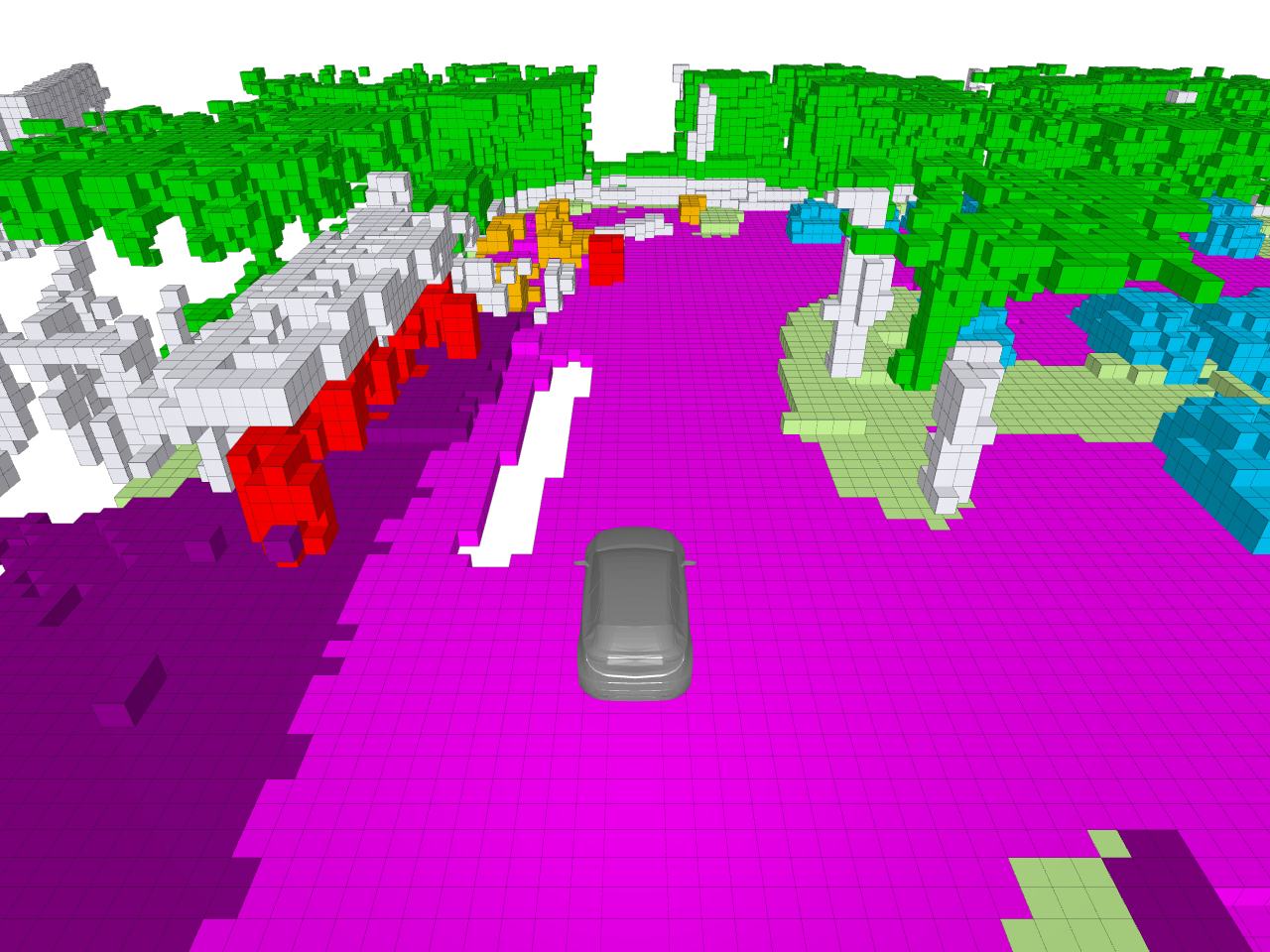}} & 
		\frame{\includegraphics[trim=0cm 0cm 0cm 1cm, clip, width=.175\textwidth]{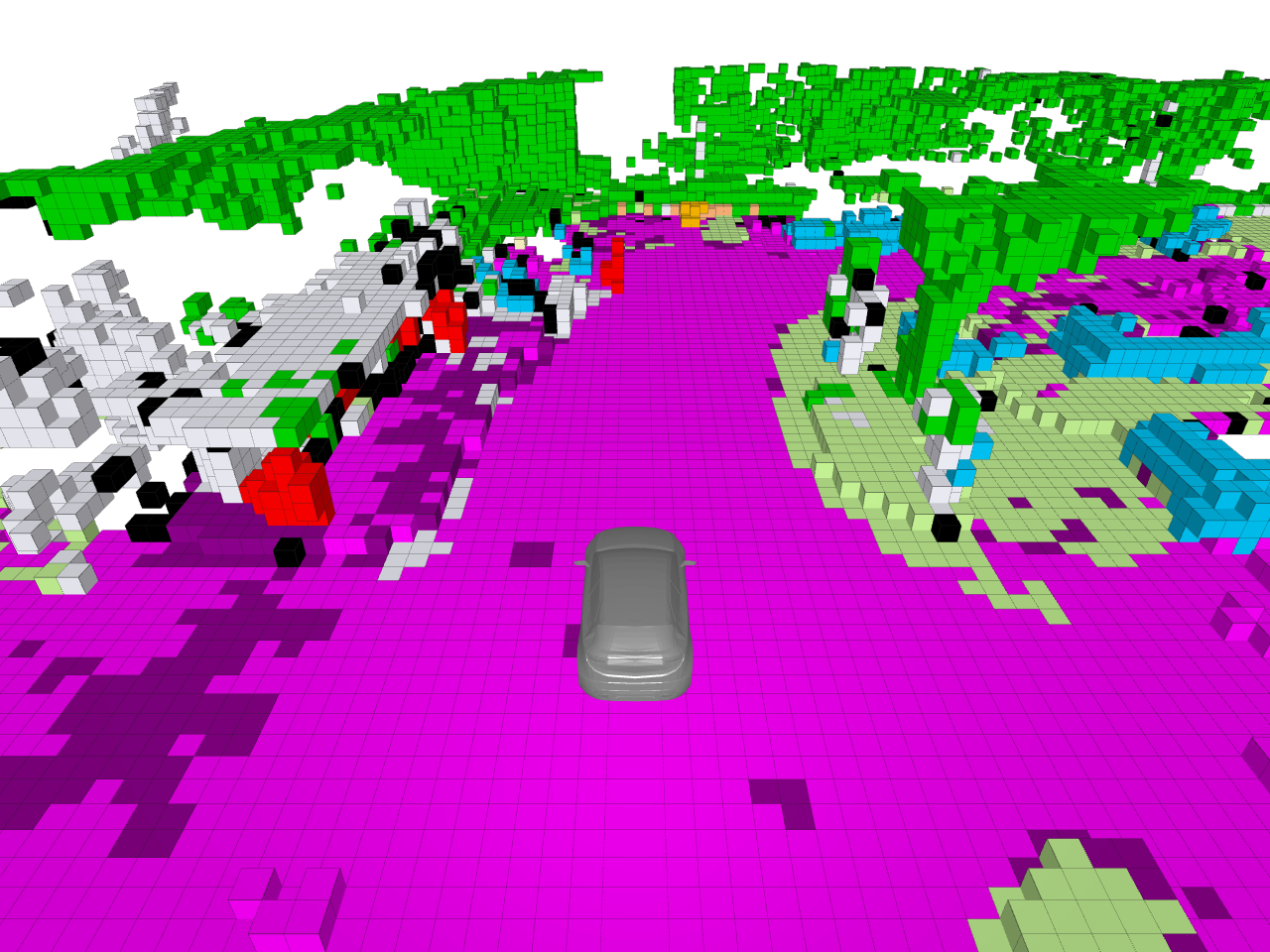}} & 
		\frame{\includegraphics[trim=0cm 0cm 0cm 1cm, clip, width=.175\textwidth]{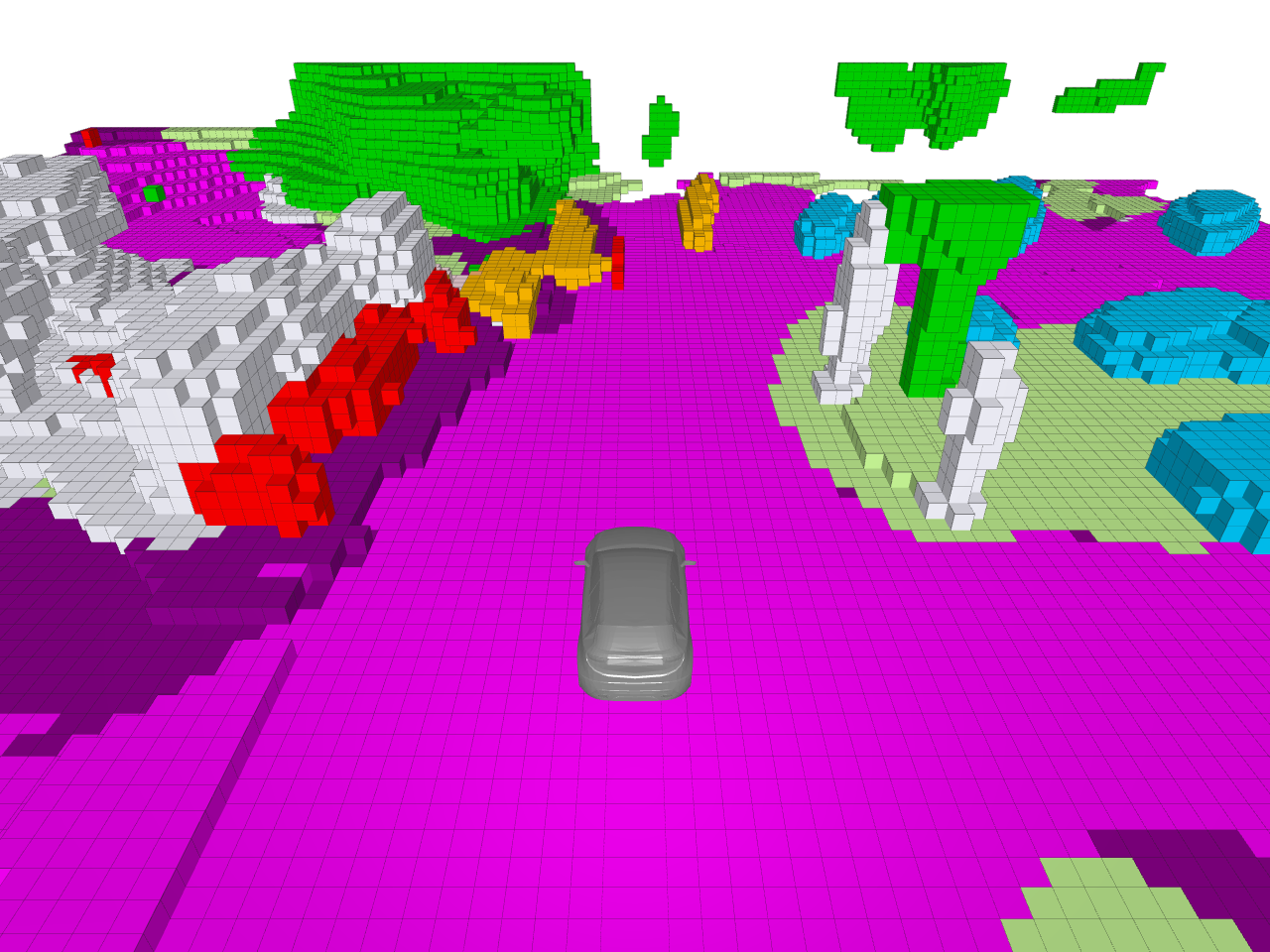}} \\

	\end{tabular}
	\caption{\textbf{Qualitative results on the Occ3D-nuScenes dataset.}
	We show the images, ground truth occupancy, the \textit{\paperName{}} pseudo-labels and the predictions of \textit{\paperName{}} + STCOcc~\cite{liao2025stcocc}.
	Best viewed when zoomed in.
	}
	\label{tab:results}
    \vspace{-2mm}
\end{figure*}

As summarized in \cref{table:main_complete}, our method establishes new state-of-the-art results across all evaluated architectures in the shelf-supervised occupancy estimation setting. 
Models trained on our \paperName{} pseudo-labels consistently outperform prior approaches, achieving improvements of up to $+5.79$ mIoU and $+9.23$ geometric IoU over the previous best-performing method, GaussianFlowOcc~\cite{Boeder_2025_ICCV}. 
This corresponds to relative gains of 34\% in mIoU and 20\% in geometric IoU. 
We believe the strong performance stems from our direct 3D supervision strategy.
High-quality pseudo-labels enable occupancy networks to be trained natively in 3D space, avoiding the instability and ambiguity associated with 2D rendering-based supervision. 
The inherently multi-view consistent geometric predictions from MapAnything~\cite{keetha2025mapanything} provide metrically accurate 3D supervision, facilitating more stable optimization and improved volumetric reasoning compared to methods relying on 2D depth rendering. 
The comparison in \cref{fig:teaser} further illustrates the clear performance gap between 2D rendering-based methods and our 3D shelf-supervised approach.
A similar trend is observed on the RayIoU metric~\cite{liu2024fully}, which provides a more depth-consistent evaluation of volumetric predictions. 
The results in \cref{table:main_rayiou} show that across all evaluated ranges (1m, 2m, and 4m), our pseudo-labels lead to consistent improvements in RayIoU when training COTR, CVT-Occ and STCOcc, highlighting the generality and robustness of the generated supervision signal.
Furthermore, the qualitative results in \cref{tab:results} demonstrate several key advantages of our approach.
The pseudo-labels generated by our framework are noticeably denser than the LiDAR-based ground truth, providing more comprehensive spatial coverage of the scene.
When training on these labels, the networks learn to effectively suppress noise and artifacts present in the raw pseudo-labels, yielding smooth and coherent occupancy maps.
Moreover, the models demonstrate strong completion capabilities, successfully reconstructing the full extent of partially visible objects and improving overall scene consistency.
We provide further results and comparisons with LiDAR-based and supervised approaches in the supplementary material.

\begin{table}
	\begin{center}
		\small
        \caption{
            \textbf{Occupancy estimation performance in terms of RayIoU~\cite{liu2024fully}.}
            We compare the performance of STCOcc trained with our \textit{ShelfOcc} labels against previous works.
        }
		\label{table:main_rayiou}
        \resizebox{\columnwidth}{!}{
    		\begin{tabular}{l|cccc}
    			\hline
    			Method & mRayIoU & RayIoU@1 & RayIoU@2 & RayIoU@4 \\
    			\hline
                GaussianOcc~\cite{gan2024gaussianocc} & 11.85 & 8.69 & 11.90 & 14.95 \\
                GaussianFlowOcc~\cite{Boeder_2025_ICCV} & 16.47 & 11.81 & 16.58 & 20.98 \\
                \hline
    			ShelfOcc + COTR~\cite{ma2024cotr} & 17.24 & 12.22 & 17.33 & 22.18 \\
    			ShelfOcc + CVT-Occ~\cite{ye2024cvtocc} & \textit{18.28} & \textit{13.11} & \textit{18.29} & \textit{23.45} \\
    			ShelfOcc + STCOcc~\cite{liao2025stcocc} & \textbf{19.97} & \textbf{14.38} & \textbf{20.07} & \textbf{25.47} \\
                \hline
    		\end{tabular}
            \addtolength{\tabcolsep}{2pt}
        }
	\end{center}
    \vspace{-4mm}
\end{table}

\subsection{Ablation Study}
\paragraph{Label Quality and Semantic Segmentation.} We first evaluate the quality of the generated pseudo-labels by directly comparing them against the Occ3D-nuScenes benchmark, as shown in \cref{table:ablation_labeleval}.
This experiment assesses the intrinsic accuracy of the pseudo-labels independent of any model training.
Without sky grounding, the pseudo-labels already capture the overall scene structure but contain substantial noise and misclassifications.
Introducing the proposed sky grounding technique leads to a notable improvement of $+3.41$ mIoU (a 55\% relative increase) and $+8.79$ geometric IoU (a 51\% relative increase), indicating that better sky segmentation substantially enhances 2D semantic mask quality and reduces false positives and negatives in the projected 3D labels.
While the pseudo-labels capture the general scene structure, their performance remains considerably below that of models trained on them, confirming that off-the-shelf foundation models alone are insufficient for high-quality occupancy prediction.
However, when used as supervision, the models trained on these pseudo-labels exhibit strong generalization and completion capabilities, filling in missing geometry and correcting inconsistencies. 
The direct gain from sky grounding persists but is smaller after training ($+0.39$ mIoU and $+1.8 $geometric IoU), indicating that the model can effectively learn to correct residual noise and resolve label ambiguities.
\begin{table}[!ht]
	\begin{center}
		\small
        \caption{
            \textbf{Evaluation of pseudo-label quality and sky grounding.}
            We report the occupancy estimation performance of the generated \textit{ShelfOcc} pseudo-labels when directly evaluated on the Occ3D-nuScenes benchmark, treating them as model predictions.
        }
		\label{table:ablation_labeleval}
		\begin{tabular}{c|c|cc}
			\hline
            w/ STCOcc~\cite{liao2025stcocc} & Sky Grounding & mIoU & IoU \\
			\hline
            \xmark & \xmark & 6.21 & 17.21 \\
            \xmark & \cmark & 9.62 & 26.00 \\
            
			\cmark & \xmark & 22.48 & 54.26 \\
			\cmark & \cmark &  22.87 & 56.14 \\
            \hline
		\end{tabular}
	\end{center}
\vspace{-5mm}
\end{table}

\paragraph{Visibility Mask.}
We analyze the influence of using the generated camera visibility mask, which restricts supervision to regions observable by the cameras, cf.~\cref{table:ablation_cameramask}.
Applying the mask leads to a significant improvement in both IoU and mIoU compared to training without it, highlighting its importance for 3D supervision in this setting.

\paragraph{Confidence Filter.}
We further ablate the effect of the proposed confidence filtering techniques in \cref{sec:filter}.
As shown in \cref{table:ablation_conf_filter}, disabling the confidence filter during pseudo-label generation leads to a decrease of $-1.88$ in mIoU and $-1.63$ in IoU, showing that our proposed pipeline produces cleaner pseudo-labels.

\begin{table}[!h]
    \begin{minipage}[t]{0.47\columnwidth}
        \centering
        \caption{
           \textbf{Ablation of camera mask.}
           Training STCOcc w/o mask, evaluated w/ mask.
           }
        \label{table:ablation_cameramask}
        \resizebox{\textwidth}{!}{
        \begin{tabular}{c|cc}
            \hline
            \phantom{ii}Camera Mask\phantom{ii} & mIoU & IoU \\
            \hline
            \xmark & 13.41 & 26.21 \\
             \cmark & 22.87 & 56.14 \\
            \hline
        \end{tabular}
        }
    \end{minipage}
    \hfill
    \begin{minipage}[t]{0.47\columnwidth}
        \centering
        \caption{
           \textbf{Ablation of conf. filter.}
           Training STCOcc on pseudo-labels w/o conf. filters.
           }
        \label{table:ablation_conf_filter}
        \resizebox{\textwidth}{!}{
        \begin{tabular}{c|cc}
            \hline
            Confidence Filter & mIoU & IoU \\
            \hline
            \xmark & 20.99 & 54.51 \\
             \cmark & 22.87 & 56.14 \\
            \hline
        \end{tabular}
        }
    \end{minipage}
    \vspace{-3mm}
\end{table}

\paragraph{Aggregation.} We further ablate the geometric aggregation pipeline to isolate the benefits of each version (see \cref{fig:teaser}).
We compare three configurations: (1) Pseudo-label generation on a per-frame basis, (2) Aggregation over time without explicit handling of dynamic objects, (3) Full pipeline including dynamic object processing.
We generate pseudo-labels with each configuration and train STCOcc on each, cf.~\cref{table:ablation_accumulation}.
Even the single-frame version surpasses previous shelf-supervised approaches in both IoU and mIoU, underscoring the inherent strength of our 3D supervision.
Temporal aggregation yields a substantial boost, particularly in geometric IoU, by reinforcing consistent static scene structures.
However, this includes practically unacceptable model violations for dynamic objects.
Finally, incorporating explicit dynamic object handling achieves an additional relative gain of approximately 5\% in mIoU, indicating that accurate modeling of dynamic objects, despite their rarity, is essential for autonomous driving.

\begin{table}
\begin{center}
	\small
	\caption{
		\textbf{Effect of scene accumulation and dynamic object handling.}
		We show results when training STCOcc on \textit{ShelfOcc} pseudo-labels and not aggregating the point clouds across multiple frames or without handling dynamic objects during accumulation.
		}
	\label{table:ablation_accumulation}
	\begin{tabular}{cc|cc}
		\hline
		  \thead{Temporal \\ Aggregation} & \thead{Dynamic \\ object handling} & mIoU & IoU \\
		\hline
		  \xmark & \xmark & 19.29 & 48.22 \\
		  \cmark & \xmark & 21.70 & 55.84 \\
		  \cmark & \cmark & 22.87 & 56.14 \\
		\hline
	\end{tabular}
\end{center}
\vspace{-3mm}
\end{table}

\section{Conclusion and Future Work} \label{sec:conclusion}
In this paper, we introduced \paperName{}, a novel training framework for shelf-supervised occupancy estimation.
We presented a comprehensive framework to generate high-quality, metrically-scaled 3D semantic voxel pseudo-labels by leveraging a 3D geometry foundation model (MapAnything~\cite{keetha2025mapanything}) in conjunction with a 2D semantic foundation model (GroundedSAM~\cite{ren2024grounded}).
Our generation pipeline addresses challenges posed by large dynamic scenes, enabling the creation of precise 3D supervisory signals.
By training different occupancy networks with these 3D pseudo-labels, we demonstrated substantial performance improvements over previous shelf-supervised methods, notably closing the gap to methods trained with full LiDAR supervision.
This work marks a significant milestone in making high-performance 3D occupancy estimation more scalable and accessible for real-world autonomous driving applications.

Future work could focus on further enhancing the quality of the generated pseudo-labels, particularly for rare or fine-grained categories.
Current models struggle with long-tail classes that are underrepresented or visually ambiguous, which can lead to incomplete or noisy semantic occupancy.
Another important avenue is the improved modeling of dynamic objects.
In the current framework, dynamic elements are reconstructed based on single-frame observations, which limits the completeness of their 3D shapes.
Integrating motion cues such as optical flow, scene flow, or multi-frame object tracking could enable temporal alignment and accumulation of dynamic object geometry across time.
This would not only produce more complete dynamic reconstructions but also extend shelf-supervised 3D labeling toward temporally consistent 4D scene understanding.

{
    \small
    \bibliographystyle{ieeenat_fullname}
    \bibliography{main}

@String(CVPR= {IEEE Conf. Comput. Vis. Pattern Recog.})

@String(ICCV= {Int. Conf. Comput. Vis.})

@String(ECCV= {Eur. Conf. Comput. Vis.})

@String(CVPR  = {CVPR})

@String(ICCV  = {ICCV})

@String(ECCV  = {ECCV})

@article{li2023fb,
    title={FB-OCC: 3D Occupancy Prediction based on Forward-Backward View Transformation},
    author={Li, Zhiqi and Yu, Zhiding and Austin, David and Fang, Mingsheng and Lan, Shiyi and Kautz, Jan and Alvarez, Jose M},
    journal={arXiv preprint arXiv:2307.01492},
    year={2023}
}

@inproceedings{tong2023scene,
  title={Scene as occupancy},
  author={Tong, Wenwen and Sima, Chonghao and Wang, Tai and Chen, Li and Wu, Silei and Deng, Hanming and Gu, Yi and Lu, Lewei and Luo, Ping and Lin, Dahua and others},
  booktitle={Proceedings of the IEEE/CVF International Conference on Computer Vision},
  pages={8406--8415},
  year={2023}
}

@article{pan2023renderocc,
  title={RenderOcc: Vision-Centric 3D Occupancy Prediction with 2D Rendering Supervision},
  author={Pan, Mingjie and Liu, Jiaming and Zhang, Renrui and Huang, Peixiang and Li, Xiaoqi and Liu, Li and Zhang, Shanghang},
  journal={arXiv preprint arXiv:2309.09502},
  year={2023}
}

@article{zhang2023occformer,
  title={OccFormer: Dual-path Transformer for Vision-based 3D Semantic Occupancy Prediction},
  author={Zhang, Yunpeng and Zhu, Zheng and Du, Dalong},
  journal={arXiv preprint arXiv:2304.05316},
  year={2023}
}

@inproceedings{cao2022monoscene,
  title={Monoscene: Monocular 3d semantic scene completion},
  author={Cao, Anh-Quan and de Charette, Raoul},
  booktitle={Proceedings of the IEEE/CVF Conference on Computer Vision and Pattern Recognition},
  pages={3991--4001},
  year={2022}
}

@inproceedings{li2023voxformer,
  title={Voxformer: Sparse voxel transformer for camera-based 3d semantic scene completion},
  author={Li, Yiming and Yu, Zhiding and Choy, Christopher and Xiao, Chaowei and Alvarez, Jose M and Fidler, Sanja and Feng, Chen and Anandkumar, Anima},
  booktitle={Proceedings of the IEEE/CVF Conference on Computer Vision and Pattern Recognition},
  pages={9087--9098},
  year={2023}
}

@inproceedings{caesar2020nuscenes,
  title={{nuScenes}: A multimodal dataset for autonomous driving},
  author={Caesar, Holger and Bankiti, Varun and Lang, Alex H and Vora, Sourabh and Liong, Venice Erin and Xu, Qiang and Krishnan, Anush and Pan, Yu and Baldan, Giancarlo and Beijbom, Oscar},
  booktitle={Proceedings of the IEEE/CVF conference on computer vision and pattern recognition},
  pages={11621--11631},
  year={2020}
}

@article{huang2022bevdet4d,
  title={BEVDet4D: Exploit Temporal Cues in Multi-camera 3D Object Detection},
  author={Huang, Junjie and Huang, Guan},
  journal={arXiv preprint arXiv:2203.17054},
  year={2022}
}

@inproceedings{huang2023tri,
  title={Tri-perspective view for vision-based 3d semantic occupancy prediction},
  author={Huang, Yuanhui and Zheng, Wenzhao and Zhang, Yunpeng and Zhou, Jie and Lu, Jiwen},
  booktitle={Proceedings of the IEEE/CVF Conference on Computer Vision and Pattern Recognition},
  pages={9223--9232},
  year={2023}
}

@article{mildenhall2021nerf,
  title={Nerf: Representing scenes as neural radiance fields for view synthesis},
  author={Mildenhall, Ben and Srinivasan, Pratul P and Tancik, Matthew and Barron, Jonathan T and Ramamoorthi, Ravi and Ng, Ren},
  journal={Communications of the ACM},
  volume={65},
  number={1},
  pages={99--106},
  year={2021},
  publisher={ACM New York, NY, USA}
}

@article{fong2022panoptic,
  title={{Panoptic nuScenes}: A large-scale benchmark for lidar panoptic segmentation and tracking},
  author={Fong, Whye Kit and Mohan, Rohit and Hurtado, Juana Valeria and Zhou, Lubing and Caesar, Holger and Beijbom, Oscar and Valada, Abhinav},
  journal={IEEE Robotics and Automation Letters},
  volume={7},
  number={2},
  pages={3795--3802},
  year={2022},
  publisher={IEEE}
}

@article{gan2023simple,
  title={A Simple Attempt for 3D Occupancy Estimation in Autonomous Driving},
  author={Gan, Wanshui and Mo, Ningkai and Xu, Hongbin and Yokoya, Naoto},
  journal={arXiv preprint arXiv:2303.10076},
  year={2023}
}

@article{zhang2023occnerf,
  title={OccNeRF: Self-Supervised Multi-Camera Occupancy Prediction with Neural Radiance Fields},
  author={Zhang, Chubin and Yan, Juncheng and Wei, Yi and Li, Jiaxin and Liu, Li and Tang, Yansong and Duan, Yueqi and Lu, Jiwen},
  journal={arXiv preprint arXiv:2312.09243},
  year={2023}
}

@article{huang2023selfocc,
  title={Selfocc: Self-supervised vision-based 3d occupancy prediction},
  author={Huang, Yuanhui and Zheng, Wenzhao and Zhang, Borui and Zhou, Jie and Lu, Jiwen},
  journal={arXiv preprint arXiv:2311.12754},
  year={2023}
}

@article{tian2023occ3d,
  title={Occ3d: A large-scale 3d occupancy prediction benchmark for autonomous driving},
  author={Tian, Xiaoyu and Jiang, Tao and Yun, Longfei and Wang, Yue and Wang, Yilun and Zhao, Hang},
  journal={arXiv preprint arXiv:2304.14365},
  year={2023}
}

@inproceedings{hayler2024s4c,
  title={S4C: Self-supervised semantic scene completion with neural fields},
  author={Hayler, Adrian and Wimbauer, Felix and Muhle, Dominik and Rupprecht, Christian and Cremers, Daniel},
  booktitle={2024 International Conference on 3D Vision (3DV)},
  pages={409--420},
  year={2024},
  organization={IEEE}
}

@inproceedings{tang2024sparseocc,
  title={Sparseocc: Rethinking sparse latent representation for vision-based semantic occupancy prediction},
  author={Tang, Pin and Wang, Zhongdao and Wang, Guoqing and Zheng, Jilai and Ren, Xiangxuan and Feng, Bailan and Ma, Chao},
  booktitle={Proceedings of the IEEE/CVF Conference on Computer Vision and Pattern Recognition},
  pages={15035--15044},
  year={2024}
}

@article{huang2024gaussianformer,
  title={GaussianFormer: Scene as Gaussians for Vision-Based 3D Semantic Occupancy Prediction},
  author={Huang, Yuanhui and Zheng, Wenzhao and Zhang, Yunpeng and Zhou, Jie and Lu, Jiwen},
  journal={arXiv preprint arXiv:2405.17429},
  year={2024}
}

@article{gan2024gaussianocc,
  title={Gaussianocc: Fully self-supervised and efficient 3d occupancy estimation with gaussian splatting},
  author={Gan, Wanshui and Liu, Fang and Xu, Hongbin and Mo, Ningkai and Yokoya, Naoto},
  journal={arXiv preprint arXiv:2408.11447},
  year={2024}
}

@article{yu2024language,
  title={Language Driven Occupancy Prediction},
  author={Yu, Zhu and Pang, Bowen and Liu, Lizhe and Zhang, Runmin and Peng, Qihao and Luo, Maochun and Yang, Sheng and Chen, Mingxia and Cao, Si-Yuan and Shen, Hui-Liang},
  journal={arXiv preprint arXiv:2411.16072},
  year={2024}
}

@inproceedings{boeder2025langocc,
  title={LangOcc: Open Vocabulary Occupancy Estimation via Volume Rendering},
  author={Boeder, Simon and Gigengack, Fabian and Risse, Benjamin},
  booktitle={2025 International Conference on 3D Vision (3DV)},
  pages={200--210},
  year={2025},
  organization={IEEE}
}

@article{tan2023ovo,
  title={Ovo: Open-vocabulary occupancy},
  author={Tan, Zhiyu and Dong, Zichao and Zhang, Cheng and Zhang, Weikun and Ji, Hang and Li, Hao},
  journal={arXiv preprint arXiv:2305.16133},
  year={2023}
}

@article{vobecky2024pop,
  title={Pop-3d: Open-vocabulary 3d occupancy prediction from images},
  author={Vobecky, Antonin and Sim{\'e}oni, Oriane and Hurych, David and Gidaris, Spyridon and Bursuc, Andrei and P{\'e}rez, Patrick and Sivic, Josef},
  journal={Advances in Neural Information Processing Systems},
  volume={36},
  year={2024}
}

@inproceedings{boeder2025occflownet,
  title={OccFlowNet: Occupancy Estimation via Differentiable Rendering and Occupancy Flow},
  author={Boeder, Simon and Risse, Benjamin},
  booktitle={2025 IEEE/CVF Winter Conference on Applications of Computer Vision (WACV)},
  pages={306--316},
  year={2025},
  organization={IEEE}
}

@inproceedings{shi2025occupancy,
  title={Occupancy as set of points},
  author={Shi, Yiang and Cheng, Tianheng and Zhang, Qian and Liu, Wenyu and Wang, Xinggang},
  booktitle={European Conference on Computer Vision},
  pages={72--87},
  year={2025},
  organization={Springer}
}

@article{yu2023flashocc,
  title={FlashOcc: Fast and Memory-Efficient Occupancy Prediction via Channel-to-Height Plugin},
  author={Yu, Zichen and Shu, Changyong and Deng, Jiajun and Lu, Kangjie and Liu, Zongdai and Yu, Jiangyong and Yang, Dawei and Li, Hui and Chen, Yan},
  journal={arXiv preprint arXiv:2311.12058},
  year={2023}
}

@misc{liu2024fully,
    title={Fully Sparse 3D Occupancy Prediction}, 
    author={Haisong Liu and Yang Chen and Haiguang Wang and Zetong Yang and Tianyu Li and Jia Zeng and Li Chen and Hongyang Li and Limin Wang},
    year={2024},
    eprint={2312.17118},
    archivePrefix={arXiv},
    primaryClass={cs.CV}
}

@article{lu2023octreeocc,
  title={OctreeOcc: Efficient and Multi-Granularity Occupancy Prediction Using Octree Queries},
  author={Lu, Yuhang and Zhu, Xinge and Wang, Tai and Ma, Yuexin},
  journal={arXiv preprint arXiv:2312.03774},
  year={2023}
}

@article{jiang2023symphonize,
  title={Symphonize 3d semantic scene completion with contextual instance queries},
  author={Jiang, Haoyi and Cheng, Tianheng and Gao, Naiyu and Zhang, Haoyang and Liu, Wenyu and Wang, Xinggang},
  journal={arXiv preprint arXiv:2306.15670},
  year={2023}
}

@article{tan2024geocc,
  title={GEOcc: Geometrically Enhanced 3D Occupancy Network with Implicit-Explicit Depth Fusion and Contextual Self-Supervision},
  author={Tan, Xin and Wu, Wenbin and Zhang, Zhiwei and Fan, Chaojie and Peng, Yong and Zhang, Zhizhong and Xie, Yuan and Ma, Lizhuang},
  journal={arXiv preprint arXiv:2405.10591},
  year={2024}
}

@InProceedings{Zhao_2024_CVPR,
    author    = {Zhao, Linqing and Xu, Xiuwei and Wang, Ziwei and Zhang, Yunpeng and Zhang, Borui and Zheng, Wenzhao and Du, Dalong and Zhou, Jie and Lu, Jiwen},
    title     = {LowRankOcc: Tensor Decomposition and Low-Rank Recovery for Vision-based 3D Semantic Occupancy Prediction},
    booktitle = {Proceedings of the IEEE/CVF Conference on Computer Vision and Pattern Recognition (CVPR)},
    month     = {June},
    year      = {2024},
    pages     = {9806-9815}
}

@inproceedings{ma2024cotr,
  title={Cotr: Compact occupancy transformer for vision-based 3d occupancy prediction},
  author={Ma, Qihang and Tan, Xin and Qu, Yanyun and Ma, Lizhuang and Zhang, Zhizhong and Xie, Yuan},
  booktitle={Proceedings of the IEEE/CVF Conference on Computer Vision and Pattern Recognition},
  pages={19936--19945},
  year={2024}
}

@inproceedings{ma2024cam4docc,
  title={Cam4docc: Benchmark for camera-only 4d occupancy forecasting in autonomous driving applications},
  author={Ma, Junyi and Chen, Xieyuanli and Huang, Jiawei and Xu, Jingyi and Luo, Zhen and Xu, Jintao and Gu, Weihao and Ai, Rui and Wang, Hesheng},
  booktitle={Proceedings of the IEEE/CVF Conference on Computer Vision and Pattern Recognition},
  pages={21486--21495},
  year={2024}
}

@inproceedings{wang2024opus,
  title={Opus: occupancy prediction using a sparse set},
  author={Wang, Jiabao and Liu, Zhaojiang and Meng, Qiang and Yan, Liujiang and Wang, Ke and Yang, Jie and Liu, Wei and Hou, Qibin and Cheng, Mingming},
  booktitle={Advances in Neural Information Processing Systems},
  year={2024}
}

@inproceedings{zhang2023simple,
  title={A simple framework for open-vocabulary segmentation and detection},
  author={Zhang, Hao and Li, Feng and Zou, Xueyan and Liu, Shilong and Li, Chunyuan and Yang, Jianwei and Zhang, Lei},
  booktitle={Proceedings of the IEEE/CVF International Conference on Computer Vision},
  pages={1020--1031},
  year={2023}
}

@article{ren2024grounded,
  title={Grounded sam: Assembling open-world models for diverse visual tasks},
  author={Ren, Tianhe and Liu, Shilong and Zeng, Ailing and Lin, Jing and Li, Kunchang and Cao, He and Chen, Jiayu and Huang, Xinyu and Chen, Yukang and Yan, Feng and others},
  journal={arXiv preprint arXiv:2401.14159},
  year={2024}
}

@inproceedings{sirko2024occfeat,
  title={OccFeat: Self-supervised Occupancy Feature Prediction for Pretraining BEV Segmentation Networks},
  author={Sirko-Galouchenko, Sophia and Boulch, Alexandre and Gidaris, Spyros and Bursuc, Andrei and Vobecky, Antonin and P{\'e}rez, Patrick and Marlet, Renaud},
  booktitle={Proceedings of the IEEE/CVF Conference on Computer Vision and Pattern Recognition},
  pages={4493--4503},
  year={2024}
}

@inproceedings{caron2021emerging,
    title        = {Emerging properties in self-supervised vision transformers},
    author       = {Caron, Mathilde and Touvron, Hugo and Misra, Ishan and J{\'e}gou, Herv{\'e} and Mairal, Julien and Bojanowski, Piotr and Joulin, Armand},
    year         = 2021,
    booktitle    = ICCV,
    pages        = {9650--9660}
}

@article{oquab2023dinov2,
    title        = {Dinov2: Learning robust visual features without supervision},
    author       = {Oquab, Maxime and Darcet, Timoth{\'e}e and Moutakanni, Th{\'e}o and Vo, Huy and Szafraniec, Marc and Khalidov, Vasil and Fernandez, Pierre and Haziza, Daniel and Massa, Francisco and El-Nouby, Alaaeldin and others},
    year         = 2023,
    journal      = {arXiv preprint arXiv:2304.07193}
}

@inproceedings{radford2021learning,
  title={Learning transferable visual models from natural language supervision},
  author={Radford, Alec and Kim, Jong Wook and Hallacy, Chris and Ramesh, Aditya and Goh, Gabriel and Agarwal, Sandhini and Sastry, Girish and Askell, Amanda and Mishkin, Pamela and Clark, Jack and others},
  booktitle={International conference on machine learning},
  pages={8748--8763},
  year={2021},
  organization={PMLR}
}

@inproceedings{zheng2025veon,
  title={VEON: Vocabulary-Enhanced Occupancy Prediction},
  author={Zheng, Jilai and Tang, Pin and Wang, Zhongdao and Wang, Guoqing and Ren, Xiangxuan and Feng, Bailan and Ma, Chao},
  booktitle={European Conference on Computer Vision},
  pages={92--108},
  year={2025},
  organization={Springer}
}

@article{jiang2024gausstr,
  title={GaussTR: Foundation Model-Aligned Gaussian Transformer for Self-Supervised 3D Spatial Understanding},
  author={Jiang, Haoyi and Liu, Liu and Cheng, Tianheng and Wang, Xinjie and Lin, Tianwei and Su, Zhizhong and Liu, Wenyu and Wang, Xinggang},
  journal={arXiv preprint arXiv:2412.13193},
  year={2024}
}

@article{sun2024gsrender,
  title={GSRender: Deduplicated Occupancy Prediction via Weakly Supervised 3D Gaussian Splatting},
  author={Sun, Qianpu and Shu, Changyong and Zhou, Sifan and Yu, Zichen and Chen, Yan and Yang, Dawei and Chun, Yuan},
  journal={arXiv preprint arXiv:2412.14579},
  year={2024}
}

@inproceedings{yin2023metric3d,
  title={Metric3d: Towards zero-shot metric 3d prediction from a single image},
  author={Yin, Wei and Zhang, Chi and Chen, Hao and Cai, Zhipeng and Yu, Gang and Wang, Kaixuan and Chen, Xiaozhi and Shen, Chunhua},
  booktitle={Proceedings of the IEEE/CVF International Conference on Computer Vision},
  pages={9043--9053},
  year={2023}
}

@article{xu2025survey,
  title={A survey on occupancy perception for autonomous driving: The information fusion perspective},
  author={Xu, Huaiyuan and Chen, Junliang and Meng, Shiyu and Wang, Yi and Chau, Lap-Pui},
  journal={Information Fusion},
  volume={114},
  pages={102671},
  year={2025},
  publisher={Elsevier}
}

@article{shi2024grid,
  title={Grid-centric traffic scenario perception for autonomous driving: A comprehensive review},
  author={Shi, Yining and Jiang, Kun and Li, Jiusi and Qian, Zelin and Wen, Junze and Yang, Mengmeng and Wang, Ke and Yang, Diange},
  journal={IEEE Transactions on Neural Networks and Learning Systems},
  year={2024},
  publisher={IEEE}
}

@article{kerbl20233d,
  title={3d gaussian splatting for real-time radiance field rendering.},
  author={Kerbl, Bernhard and Kopanas, Georgios and Leimk{\"u}hler, Thomas and Drettakis, George},
  journal={ACM Trans. Graph.},
  volume={42},
  number={4},
  pages={139--1},
  year={2023}
}

@inproceedings{wu20244d,
  title={4d gaussian splatting for real-time dynamic scene rendering},
  author={Wu, Guanjun and Yi, Taoran and Fang, Jiemin and Xie, Lingxi and Zhang, Xiaopeng and Wei, Wei and Liu, Wenyu and Tian, Qi and Wang, Xinggang},
  booktitle={Proceedings of the IEEE/CVF Conference on Computer Vision and Pattern Recognition},
  pages={20310--20320},
  year={2024}
}

@inproceedings{he2016deep,
  title={Deep residual learning for image recognition},
  author={He, Kaiming and Zhang, Xiangyu and Ren, Shaoqing and Sun, Jian},
  booktitle={Proceedings of the IEEE conference on computer vision and pattern recognition},
  pages={770--778},
  year={2016}
}

@article{lu2024drivingrecon,
  title={DrivingRecon: Large 4D Gaussian Reconstruction Model For Autonomous Driving},
  author={Lu, Hao and Xu, Tianshuo and Zheng, Wenzhao and Zhang, Yunpeng and Zhan, Wei and Du, Dalong and Tomizuka, Masayoshi and Keutzer, Kurt and Chen, Yingcong},
  journal={arXiv preprint arXiv:2412.09043},
  year={2024}
}

@inproceedings{yan2024street,
  title={Street gaussians: Modeling dynamic urban scenes with gaussian splatting},
  author={Yan, Yunzhi and Lin, Haotong and Zhou, Chenxu and Wang, Weijie and Sun, Haiyang and Zhan, Kun and Lang, Xianpeng and Zhou, Xiaowei and Peng, Sida},
  booktitle={European Conference on Computer Vision},
  pages={156--173},
  year={2024},
  organization={Springer}
}

@InProceedings{Boeder_2025_ICCV,
    author    = {Boeder, Simon and Gigengack, Fabian and Risse, Benjamin},
    title     = {GaussianFlowOcc: Sparse and Weakly Supervised Occupancy Estimation using Gaussian Splatting and Temporal Flow},
    booktitle = {Proceedings of the IEEE/CVF International Conference on Computer Vision (ICCV)},
    month     = {October},
    year      = {2025},
    pages     = {24943-24954}
}

@article{keetha2025mapanything,
  title={MapAnything: Universal Feed-Forward Metric 3D Reconstruction},
  author={Keetha, Nikhil and M{\"u}ller, Norman and Sch{\"o}nberger, Johannes and Porzi, Lorenzo and Zhang, Yuchen and Fischer, Tobias and Knapitsch, Arno and Zauss, Duncan and Weber, Ethan and Antunes, Nelson and others},
  journal={arXiv preprint arXiv:2509.13414},
  year={2025}
}

@inproceedings{wang2025vggt,
  title={Vggt: Visual geometry grounded transformer},
  author={Wang, Jianyuan and Chen, Minghao and Karaev, Nikita and Vedaldi, Andrea and Rupprecht, Christian and Novotny, David},
  booktitle={Proceedings of the Computer Vision and Pattern Recognition Conference},
  pages={5294--5306},
  year={2025}
}

@inproceedings{liu2024grounding,
  title={Grounding dino: Marrying dino with grounded pre-training for open-set object detection},
  author={Liu, Shilong and Zeng, Zhaoyang and Ren, Tianhe and Li, Feng and Zhang, Hao and Yang, Jie and Jiang, Qing and Li, Chunyuan and Yang, Jianwei and Su, Hang and others},
  booktitle={European conference on computer vision},
  pages={38--55},
  year={2024},
  organization={Springer}
}

@article{kirillov2023segany,
  title={Segment Anything},
  author={Kirillov, Alexander and Mintun, Eric and Ravi, Nikhila and Mao, Hanzi and Rolland, Chloe and Gustafson, Laura and Xiao, Tete and Whitehead, Spencer and Berg, Alexander C. and Lo, Wan-Yen and Doll{\'a}r, Piotr and Girshick, Ross},
  journal={arXiv:2304.02643},
  year={2023}
}

@misc{hayes2025easyocc,
      title={EasyOcc: 3D Pseudo-Label Supervision for Fully Self-Supervised Semantic Occupancy Prediction Models}, 
      author={Seamie Hayes and Ganesh Sistu and Ciarán Eising},
      year={2025},
      eprint={2509.26087},
      archivePrefix={arXiv},
      primaryClass={cs.CV},
      url={https://arxiv.org/abs/2509.26087}, 
}

@article{li2025ago,
  title={AGO: Adaptive Grounding for Open World 3D Occupancy Prediction},
  author={Li, Peizheng and Ding, Shuxiao and Zhou, You and Zhang, Qingwen and Inak, Onat and Triess, Larissa and Hanselmann, Niklas and Cordts, Marius and Zell, Andreas},
  journal={arXiv preprint arXiv:2504.10117},
  year={2025}
}

@inproceedings{liao2025stcocc,
  title={STCOcc: Sparse spatial-temporal cascade renovation for 3d occupancy and scene flow prediction},
  author={Liao, Zhimin and Wei, Ping and Chen, Shuaijia and Wang, Haoxuan and Ren, Ziyang},
  booktitle={Proceedings of the Computer Vision and Pattern Recognition Conference},
  pages={1516--1526},
  year={2025}
}

@article{zhou2025autoocc,
  title={AutoOcc: Automatic Open-Ended Semantic Occupancy Annotation via Vision-Language Guided Gaussian Splatting},
  author={Zhou, Xiaoyu and Wang, Jingqi and Wang, Yongtao and Wei, Yufei and Dong, Nan and Yang, Ming-Hsuan},
  journal={arXiv preprint arXiv:2502.04981},
  year={2025}
}

@article{ye2024cvtocc,
    title={CVT-Occ: Cost Volume Temporal Fusion for 3D Occupancy Prediction},
    author={Ye, Zhangchen and Jiang, Tao and Xu, Chenfeng and Li, Yiming and Zhao, Hang},
    journal={arXiv preprint arXiv:2409.13430},
    year={2024},
    url={https://arxiv.org/abs/2409.13430}
}

@article{tempfli2025vespa,
  title={VESPA: Towards un (Human) supervised Open-World Pointcloud Labeling for Autonomous Driving},
  author={Tempfli, Levente and Rivera, Esteban and Lienkamp, Markus},
  journal={arXiv preprint arXiv:2507.20397},
  year={2025}
}

@inproceedings{feng2025gaussian,
  title={Gaussian-based World Model: Gaussian Priors for Voxel-Based Occupancy Prediction and Future Motion Prediction},
  author={Feng, Tuo and Wang, Wenguan and Yang, Yi},
  booktitle={Proceedings of the IEEE/CVF International Conference on Computer Vision},
  pages={25239--25249},
  year={2025}
}

@article{wang2024occsora,
  title={Occsora: 4d occupancy generation models as world simulators for autonomous driving},
  author={Wang, Lening and Zheng, Wenzhao and Ren, Yilong and Jiang, Han and Cui, Zhiyong and Yu, Haiyang and Lu, Jiwen},
  journal={arXiv preprint arXiv:2405.20337},
  year={2024}
}

@inproceedings{zheng2024occworld,
  title={Occworld: Learning a 3d occupancy world model for autonomous driving},
  author={Zheng, Wenzhao and Chen, Weiliang and Huang, Yuanhui and Zhang, Borui and Duan, Yueqi and Lu, Jiwen},
  booktitle={European conference on computer vision},
  pages={55--72},
  year={2024},
  organization={Springer}
}

@inproceedings{zhou2022detecting,
  title={Detecting Twenty-thousand Classes using Image-level Supervision},
  author={Zhou, Xingyi and Girdhar, Rohit and Joulin, Armand and Kr{\"a}henb{\"u}hl, Philipp and Misra, Ishan},
  booktitle={ECCV},
  year={2022}
}

@article{lan2024text4seg,
  title={Text4seg: Reimagining image segmentation as text generation},
  author={Lan, Mengcheng and Chen, Chaofeng and Zhou, Yue and Xu, Jiaxing and Ke, Yiping and Wang, Xinjiang and Feng, Litong and Zhang, Wayne},
  journal={arXiv preprint arXiv:2410.09855},
  year={2024}
}

@inproceedings{wang2024dust3r,
  title={Dust3r: Geometric 3d vision made easy},
  author={Wang, Shuzhe and Leroy, Vincent and Cabon, Yohann and Chidlovskii, Boris and Revaud, Jerome},
  booktitle={Proceedings of the IEEE/CVF Conference on Computer Vision and Pattern Recognition},
  pages={20697--20709},
  year={2024}
}

@inproceedings{leroy2024grounding,
  title={Grounding image matching in 3d with mast3r},
  author={Leroy, Vincent and Cabon, Yohann and Revaud, J{\'e}r{\^o}me},
  booktitle={European Conference on Computer Vision},
  pages={71--91},
  year={2024},
  organization={Springer}
}

@inproceedings{chen2025alocc,
  title={ALOcc: Adaptive Lifting-Based 3D Semantic Occupancy and Cost Volume-Based Flow Predictions},
  author={Chen, Dubing and Fang, Jin and Han, Wencheng and Cheng, Xinjing and Yin, Junbo and Xu, Chengzhong and Khan, Fahad Shahbaz and Shen, Jianbing},
  booktitle={Proceedings of the IEEE/CVF International Conference on Computer Vision},
  pages={4156--4166},
  year={2025}
}

@INPROCEEDINGS{9577840,
  author={Ye, Yufei and Tulsiani, Shubham and Gupta, Abhinav},
  booktitle={2021 IEEE/CVF Conference on Computer Vision and Pattern Recognition (CVPR)}, 
  title={Shelf-Supervised Mesh Prediction in the Wild}, 
  year={2021},
  volume={},
  number={},
  pages={8839-8848},
  doi={10.1109/CVPR46437.2021.00873}}

@article{khurana2024shelf,
  title={Shelf-Supervised Cross-Modal Pre-Training for 3D Object Detection},
  author={Khurana, Mehar and Peri, Neehar and Hays, James and Ramanan, Deva},
  journal={arXiv preprint arXiv:2406.10115},
  year={2024}
}

@INPROCEEDINGS{imagenet,
  author={Deng, Jia and Dong, Wei and Socher, Richard and Li, Li-Jia and Kai Li and Li Fei-Fei},
  booktitle={2009 IEEE Conference on Computer Vision and Pattern Recognition}, 
  title={ImageNet: A large-scale hierarchical image database}, 
  year={2009},
  volume={},
  number={},
  pages={248-255},
  doi={10.1109/CVPR.2009.5206848}}

@InProceedings{Ye_2025_ICCV,
    author    = {Ye, Baijun and Qin, Minghui and Zhang, Saining and Gong, Moonjun and Zhu, Shaoting and Zhao, Hao and Zhao, Hang},
    title     = {GS-Occ3D: Scaling Vision-only Occupancy Reconstruction with Gaussian Splatting},
    booktitle = {Proceedings of the IEEE/CVF International Conference on Computer Vision (ICCV)},
    month     = {October},
    year      = {2025},
    pages     = {25925-25937}
}

@InProceedings{Chen_2025_ICCV,
    author    = {Chen, Dubing and Zheng, Huan and Zhou, Yucheng and Li, Xianfei and Liao, Wenlong and He, Tao and Peng, Pai and Shen, Jianbing},
    title     = {Semantic Causality-Aware Vision-Based 3D Occupancy Prediction},
    booktitle = {Proceedings of the IEEE/CVF International Conference on Computer Vision (ICCV)},
    month     = {October},
    year      = {2025},
    pages     = {24878-24888}
}

@article{gebraad2025leap,
  title={LeAP: Consistent multi-domain 3D labeling using Foundation Models},
  author={Gebraad, Simon and Palffy, Andras and Caesar, Holger},
  journal={arXiv preprint arXiv:2502.03901},
  year={2025}
}

@inproceedings{song2025coda,
  title={Coda-4dgs: Dynamic gaussian splatting with context and deformation awareness for autonomous driving},
  author={Song, Rui and Liang, Chenwei and Xia, Yan and Zimmer, Walter and Cao, Hu and Caesar, Holger and Festag, Andreas and Knoll, Alois},
  booktitle={Proceedings of the IEEE/CVF International Conference on Computer Vision},
  pages={28031--28041},
  year={2025}
}

@article{huang2024textit,
  title={S3Gaussian: Self-Supervised Street Gaussians for Autonomous Driving},
  author={Huang, Nan and Wei, Xiaobao and Zheng, Wenzhao and An, Pengju and Lu, Ming and Zhan, Wei and Tomizuka, Masayoshi and Keutzer, Kurt and Zhang, Shanghang},
  journal={arXiv preprint arXiv:2405.20323},
  year={2024}
}
}

\clearpage
\setcounter{page}{1}
\maketitlesupplementary
\appendix
\renewcommand\thefigure{\thesection.\arabic{figure}}
\setcounter{figure}{0}

\renewcommand\thetable{\thesection.\arabic{table}}
\setcounter{table}{0}

\section{Additional Qualitative Results}

\begin{figure*}[!htbp]
    \centering
    \includegraphics[page=2, width=\textwidth]{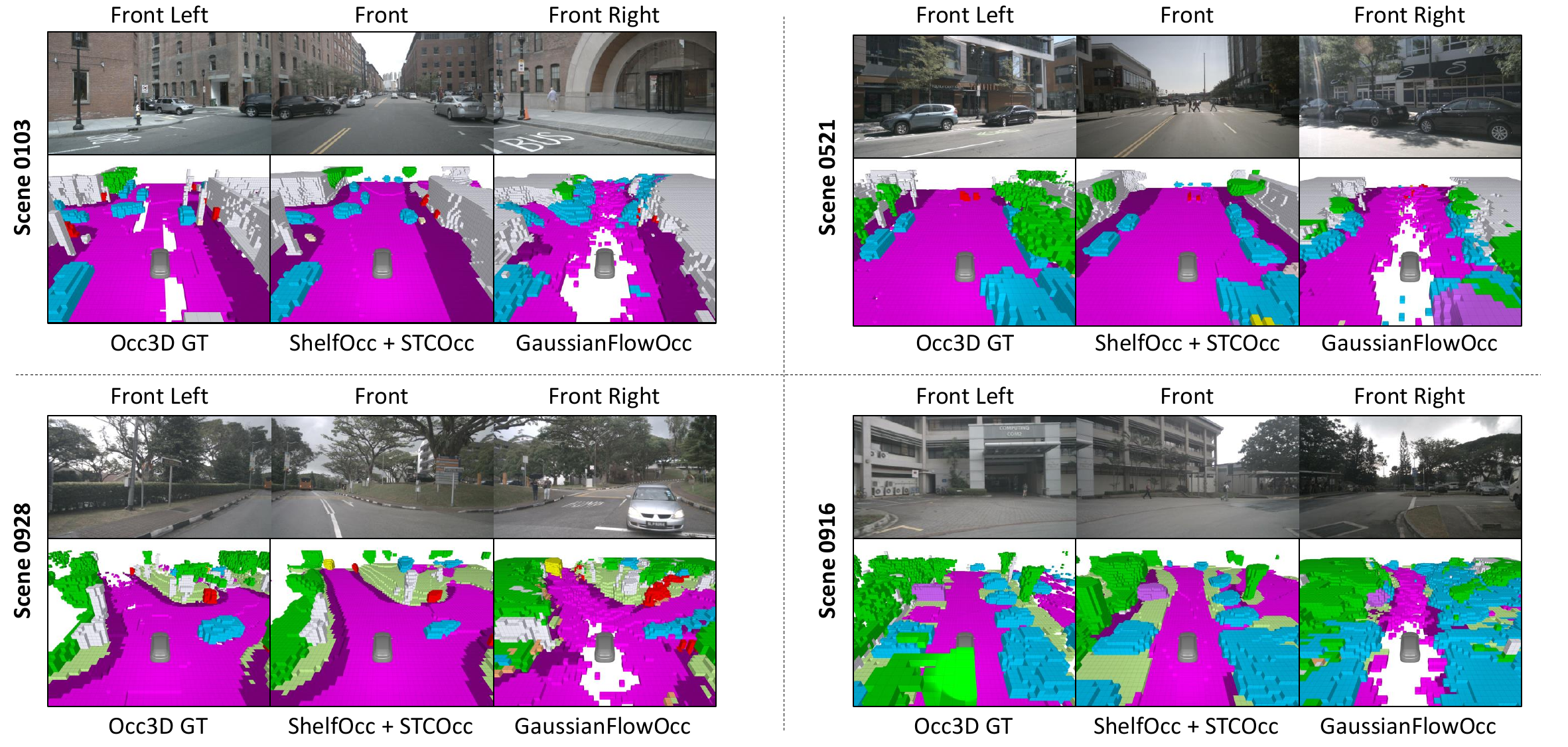}
    \caption{
    \textbf{Qualitative comparison with previous state-of-the-art.}
    We show predictions from STCOcc~\cite{liao2025stcocc} trained on our ShelfOcc pseudo-labels, compared against GaussianFlowOcc~\cite{Boeder_2025_ICCV} and the Occ3D-nuScenes ground truth.
    STCOcc produces cleaner and more geometrically consistent occupancy predictions, demonstrating the benefits of our 3D supervision.}
    \label{fig:qual_comparison}
\end{figure*}

We provide further qualitative comparisons to complement the results in the main paper. 
First, we compare STCOcc~\cite{liao2025stcocc} trained on our pseudo-labels against the previous state-of-the-art method GaussianFlowOcc~\cite{Boeder_2025_ICCV}, as well as the Occ3D-nuScenes ground truth~\cite{tian2023occ3d} in \cref{fig:qual_comparison}.
For each scene, we show the three front-facing camera images alongside 3D predictions rendered from an elevated third-person viewpoint behind the ego vehicle for a single frame.
STCOcc produces clean, dense, and well-regularized occupancy predictions with minimal noise or depth bleeding, whereas GaussianFlowOcc exhibits pronounced artifacts stemming from its 2D supervision pipeline.
It is clearly visible that the model trained with our framework can correctly estimate the 3D shape of objects, while previous work suffers from depth bleeding.

\begin{figure*}[!htbp]
    \centering
    \includegraphics[page=3, trim=0cm 0cm 3.4cm 0cm, clip, width=\textwidth]{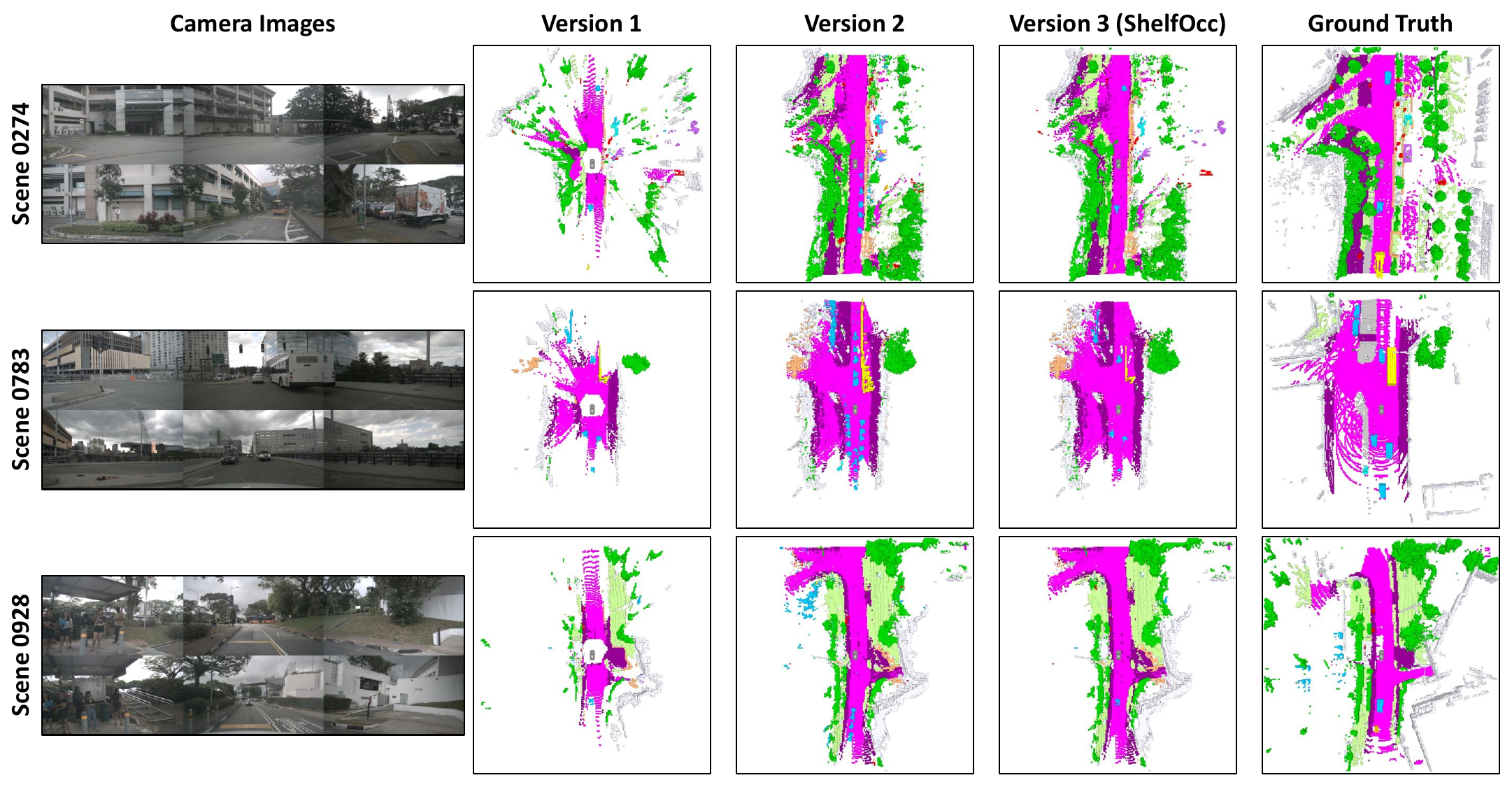}
    \caption{
    \textbf{Qualitative comparison of the different \textit{versions} of our proposed pipeline.}
    We visualize pseudo-labels produced by the three pipeline variants introduced in the main paper:
    (1) the naïve single-frame approach,
    (2) full temporal aggregation without handling motion, and
    (3) our final design, which aggregates static geometry while treating dynamic objects separately.
    The comparison highlights how version 3 avoids sparsity, object trails, and missing objects, resulting in clean and coherent 3D supervision.}
    \label{fig:qual_versions}
\end{figure*}

We also visualize the effect of the different \emph{versions} of our proposed pipeline introduced in \cref{fig:teaser}, rendered from a top-down viewpoint in \cref{fig:qual_versions}.
The naïve single-frame variant (version~1) yields sparse and incomplete geometry, missing large portions of the scene.
Aggregating all points across the sequence without distinguishing motion (version~2) introduces object trails and leads to missing objects when low-confidence points are filtered out, both of which degrade the supervision quality.
In contrast, our final design (version~3), which explicitly separates static and dynamic content and applies confidence filtering only to the static scene, produces a dense, coherent scene without trails or missing objects.

\section{Additional Quantitative Results}
\subsection{Comparison to LiDAR-Supervised Methods}
\Cref{table:main_complete_lidar} provides an extended comparison between our approach and methods that rely on LiDAR for supervision, including weakly supervised approaches AGO~\cite{li2025ago} and VEON~\cite{zheng2025veon} using LiDAR data, as well as fully supervised methods trained with semantic 3D ground truth.
Interestingly, both COTR and CVT-Occ, when trained solely on our pseudo-labels, achieve competitive performance relative to AGO and even surpass VEON, despite both of the latter relying on LiDAR for geometric supervision. 
STCOcc surpasses them even more clearly in terms of mIoU. 
Unfortunately, the authors of these methods do not report geometric IoU, where we would expect them to perform more strongly due to their access to LiDAR depth.
These findings highlight that our LiDAR-free, shelf-supervised framework can match or even outperform prior methods that depend on LiDAR supervision for geometry.
At the same time, there remains a performance gap compared to fully supervised methods trained directly on densely annotated LiDAR voxel labels.
While LiDAR-based occupancy estimation is not the focus of this work, we provide these numbers to contextualize the remaining room for improvement relative to full 3D supervision.
We omit AutoOcc~\cite{zhou2025autoocc} from this comparison, as we were unable to retrace their differing evaluation protocol.

\begin{table*}
	\begin{center}
		\caption{
			\textbf{Performance on the Occ3D-nuScenes validation set compared to methods trained with LiDAR data.}
            The \emph{LiDAR} column indicates whether a method uses raw 3D LiDAR points during training, while the \emph{Annotations} column denotes methods that rely on semantically annotated LiDAR ground truth (e.g., voxel labels from Occ3D-nuScenes).
            Despite using only camera images for supervision, our shelf-supervised pipeline outperforms prior methods that depend on LiDAR-based geometric supervision.}
		\label{table:main_complete_lidar}

		\resizebox{\textwidth}{!}{%
			\addtolength{\tabcolsep}{2pt}
			\begin{tabular}{l|cc|cc|ccccccccccccccc}
				\hline
				\noalign{\smallskip}
				Method & LiDAR & Annotations & mIoU & IoU & 
                \rotatebox{90}{\textcolor{barrier}{$\blacksquare$} barrier} &
                \rotatebox{90}{\textcolor{bicycle}{$\blacksquare$} bicycle} &
                \rotatebox{90}{\textcolor{bus}{$\blacksquare$} bus} &
                \rotatebox{90}{\textcolor{car}{$\blacksquare$} car} &
                \rotatebox{90}{\textcolor{construction}{$\blacksquare$} cons. vehicle} &
                \rotatebox{90}{\textcolor{motorcycle}{$\blacksquare$} motorcycle} &
                \rotatebox{90}{\textcolor{pedestrian}{$\blacksquare$} pedestrian} &
                \rotatebox{90}{\textcolor{cone}{$\blacksquare$} traffic cone} &
                \rotatebox{90}{\textcolor{trailer}{$\blacksquare$} trailer} &
                \rotatebox{90}{\textcolor{truck}{$\blacksquare$} truck} &
                \rotatebox{90}{\textcolor{driveable}{$\blacksquare$} driv. surf.} &
                \rotatebox{90}{\textcolor{sidewalk}{$\blacksquare$} sidewalk} &
                \rotatebox{90}{\textcolor{terrain}{$\blacksquare$} terrain} &
                \rotatebox{90}{\textcolor{manmade}{$\blacksquare$} manmade} &
                \rotatebox{90}{\textcolor{vegetation}{$\blacksquare$} vegetation}\\
				
				\noalign{\smallskip}
				\hline
				\noalign{\smallskip}
				
				AGO \cite{li2025ago} & \cmark & \xmark & 21.39 & - & 6.75 & 6.43 & 14.00 & 22.82 & 5.57 & 16.66 & 13.20 & 6.80 & 10.53 & 15.89 & 71.48 & 34.48 & 41.37 & 29.33 & 25.66 \\
				VEON \cite{li2025ago} & \cmark & \xmark & 17.07 & - & 10.40 & 6.20 & 17.70 & 12.70 & 8.50 & 7.60 & 6.50 & 5.50 & 8.20 & 11.80 & 54.50 & 25.50 & 30.20 & 25.40 & 25.40  \\

				\noalign{\smallskip}
				\hline
				\noalign{\smallskip}
                
                CVT-Occ \cite{ye2024cvtocc} & \cmark & \cmark & 42.36 & - & 49.46 & 23.57 & 49.18 & 55.63 & 23.10 & 27.85 & 28.88 & 29.07 & 34.97 & 40.98 & 81.44 & 51.37 & 54.25 & 45.94 & 39.71 \\
                COTR \cite{ma2024cotr} & \cmark & \cmark & 46.41 & 75.01 & 52.11 & 31.95 & 46.03 & 55.63 & 32.57 & 32.78 & 30.35 & 34.09 & 37.72 & 41.84 & 84.48 & 57.55 & 60.67 & 51.99 & 46.33 \\
				STCOcc \cite{liao2025stcocc} & \cmark & \cmark & 46.83 & - & 52.3 & 32.2 & 50.5 & 56.5 & 31.7 & 33.9 & 33.4 & 33.8 & 38.9 & 44.9 & 83.9 & 57.1 & 60.1 & 50.6 & 42.7 \\

				\noalign{\smallskip}
				\hline
				\noalign{\smallskip}
                
                Ours: \paperName{} + COTR \cite{ma2024cotr} & \xmark & \xmark & 18.65 & 53.71 & 9.10 & 6.20 & 22.92 & 22.08 & 1.66 & 5.94 & 9.92 & 8.55 & 0.0 & 15.32 & 67.93 & 31.13 & 38.76 & 23.11 & 17.15 \\
				
				Ours: \paperName{} + CVT-Occ \cite{ye2024cvtocc} & \xmark & \xmark & 19.21 & 52.72 & 11.53 & 6.38 & 20.39 & 21.92 & 4.20 & 10.18 & 9.02 & 10.67 & 0.89 & 13.08 & 68.42 & 31.23 & 41.42 & 22.74 & 16.15 \\
								
				Ours: \paperName{} + STCOcc \cite{liao2025stcocc} & \xmark & \xmark & 22.87 & 56.14 & 13.98 & 11.36 & 25.27 & 25.80 & 7.25 & 16.61 & 12.91 & 13.42 & 5.37 & 17.15 & 68.01 & 34.66 & 42.73 & 25.63 & 22.89 \\
				
				\noalign{\smallskip}
				\hline
			\end{tabular}
			\addtolength{\tabcolsep}{2pt}
		}
	\end{center}
\end{table*}

\subsection{Per-Class Semantic Segmentation Ablation}
\label{sec:additional_semseg}
We additionally report per-class performance for the semantic segmentation ablation (cf.~\cref{table:ablation_labeleval}) in \cref{table:main_complete_semseg}.
The results confirm that the proposed sky grounding technique substantially improves the quality of the pseudo-labels across almost all classes.
By reducing spurious false positives from the open-vocabulary detector the resulting labels become significantly cleaner and more stable.
Notably, improvements are pronounced for low-frequency classes such as \emph{bus}, \emph{traffic cone}, and \emph{truck}.
For these categories, sky grounding prevents the detector from erroneously predicting object boxes in every frame, enabling more accurate class assignments and reducing confusion with the background.

\begin{table*}
	\begin{center}
		\caption{
			\textbf{Effect of improved semantic segmentation.}
            We train STCOcc~\cite{liao2025stcocc} on our pseudo-labels with and without using the sky grounding technique. 
            Using the improved semantic segmentation also improves downstream occupancy estimation performance.
        }
		\label{table:main_complete_semseg}

		\resizebox{\textwidth}{!}{%
			\addtolength{\tabcolsep}{2pt}
			\begin{tabular}{l|c|cc|ccccccccccccccc}
				\hline
				\noalign{\smallskip}
				Method & Sky Grounding & mIoU & IoU & 
                \rotatebox{90}{\textcolor{barrier}{$\blacksquare$} barrier} &
                \rotatebox{90}{\textcolor{bicycle}{$\blacksquare$} bicycle} &
                \rotatebox{90}{\textcolor{bus}{$\blacksquare$} bus} &
                \rotatebox{90}{\textcolor{car}{$\blacksquare$} car} &
                \rotatebox{90}{\textcolor{construction}{$\blacksquare$} cons. vehicle} &
                \rotatebox{90}{\textcolor{motorcycle}{$\blacksquare$} motorcycle} &
                \rotatebox{90}{\textcolor{pedestrian}{$\blacksquare$} pedestrian} &
                \rotatebox{90}{\textcolor{cone}{$\blacksquare$} traffic cone} &
                \rotatebox{90}{\textcolor{trailer}{$\blacksquare$} trailer} &
                \rotatebox{90}{\textcolor{truck}{$\blacksquare$} truck} &
                \rotatebox{90}{\textcolor{driveable}{$\blacksquare$} driv. surf.} &
                \rotatebox{90}{\textcolor{sidewalk}{$\blacksquare$} sidewalk} &
                \rotatebox{90}{\textcolor{terrain}{$\blacksquare$} terrain} &
                \rotatebox{90}{\textcolor{manmade}{$\blacksquare$} manmade} &
                \rotatebox{90}{\textcolor{vegetation}{$\blacksquare$} vegetation}\\
				
				\noalign{\smallskip}
				\hline
				\noalign{\smallskip}
                Ours: \textit{\paperName{}} & \xmark & 6.21 & 17.21 & 3.7 & 1.93 & 3.81 & 5.57 & 1.55 & 3.27 & 3.95 & 5.39 & 0.10 & 2.87 & 20.17 & 9.27 & 10.46 & 8.21 & 12.91 \\
                Ours: \textit{\paperName{}} & \cmark & 9.62 & 26.00 & 6.58 & 3.28 & 7.02 & 8.81 & 2.57 & 4.74 & 4.97 & 8.1 & 0.12 & 5.41 & 34.59 & 14.58 & 18.16 & 10.91 & 14.45\\
                
				\noalign{\smallskip}
				\hline
				\noalign{\smallskip}
				
				Ours: \textit{\paperName{} + STCOcc \cite{liao2025stcocc}} & \xmark & 22.48 & 54.26 & 11.96 & 9.89 & 23.64 & 25.88 & 7.89 & 15.24 & 13.31 & 14.43 & 7.73 & 16.84 & 67.64 & 35.51 & 40.13 & 24.43 & 22.67 \\
                
				Ours: \textit{\paperName{} + STCOcc \cite{liao2025stcocc}} & \cmark & 22.87 & 56.14 & 13.98 & 11.36 & 25.27 & 25.80 & 7.25 & 16.61 & 12.91 & 13.42 & 5.37 & 17.15 & 68.01 & 34.66 & 42.73 & 25.63 & 22.89 \\
				
				\noalign{\smallskip}
				\hline
			\end{tabular}
			\addtolength{\tabcolsep}{2pt}
		}
	\end{center}
\end{table*}

\subsection{Ablation on Resolution and Backbone}
We further investigate the impact of input image resolution and backbone capacity on models trained with our \paperName{} pseudo-labels in \cref{table:ablation_backbone}.
For this study, we use CVT-Occ as the representative architecture.
Our results show that CVT-Occ benefits noticeably from scaling both the backbone and the input resolution. Doubling the resolution to $512 \times 1408$ and replacing the ResNet-50 encoder with a larger ResNet-101 already yields a clear improvement in semantic mIoU.
Increasing the resolution further to the full nuScenes input size leads to additional gains, improving not only mIoU but also geometric IoU.
We also experimented with VoVNet-99, which has a parameter count comparable to ResNet-101.
While its semantic mIoU is similar to that of ResNet-101, it achieves slightly lower geometric IoU, suggesting that encoder architecture plays a nontrivial role in exploiting the pseudo-label supervision.
Overall, these findings indicate that our 3D supervision pipeline can effectively leverage higher-resolution inputs and stronger backbones, offering additional room for performance scaling.

\begin{table}
\begin{center}
	\small
	\caption{
		\textbf{Ablation of Image Backbones for CVT-Occ.}
		Comparison of ResNet variants and VoVNet while simultaneously scaling image resolution along with backbone size. 
        We observe further performance increases when scaling up the model size and image resolution.
		}
	\label{table:ablation_backbone}
	\begin{tabular}{l|c|cc}
		\hline
		Backbone & Image Size & mIoU & IoU \\
		\hline
		ResNet-50 & 256 x 704 & 19.21 & 52.72 \\
		ResNet-101 & 512 x 1408 & 19.78 & 52.77 \\
		ResNet-101 & 928 x 1600 & 20.24 & 53.02 \\
		VoVNet-99 & 928 x 1600 & 20.23 & 51.84 \\
		\hline
	\end{tabular}
\end{center}
\end{table}

\section{Details on Semantic Segmentation}
We provide the full vocabulary used to query the open-vocabulary detector Grounding DINO~\cite{liu2024grounding} in \cref{table:vocabulary}.
As described in the main paper, each category is queried individually by forwarding a prompt of the form \textit{“QUERY . sky”} through the model.
Including the background token \emph{sky} encourages the detector to identify sky regions explicitly, which in turn reduces false positives for the target query.
For each forward pass, we discard all predicted boxes corresponding to \emph{sky} and retain only the boxes associated with the target query.
To ensure high-quality detections, we further filter out any box whose predicted logit falls below $0.2$. 
All remaining boxes across all query categories are aggregated and passed to the SAM segmentation model, which generates a mask for each box.
The logit of the originating Grounding DINO detection is assigned to every pixel within the corresponding SAM mask.
To construct the final semantic segmentation map, we overlay all predicted masks and perform per-pixel selection based on the highest associated logit.
This produces a dense, open-vocabulary segmentation result that serves as the semantic input to our pseudo-label generation pipeline.

\begin{table*}
\small
\begin{center}
 \caption{
    \textbf{Vocabulary used for querying Grounding DINO~\cite{liu2024grounding} for open-vocabulary object detection.}
    The left column lists the class labels, and the right column contains the prompts used during mask generation.}
    \label{table:vocabulary}    
        \begin{tabular}{l|l}
            \hline
            Class & Prompts \\
            \hline
            'barrier' & 'barricade', 'barrier' \\
            'bicycle' & 'bicycle' \\
            'bus' & 'bus' \\
            'car' & 'car', 'sedan', 'van' \\
            'construction\_vehicle' & 'excavator', 'crane' \\
            'motorcycle' & 'motorcycle', 'scooter' \\
            'pedestrian' & 'person', 'pedestrian' \\
            'traffic\_cone' & 'traffic-cone' \\
            'trailer' & 'trailer' \\
            'truck' & 'lorry', 'truck' \\
            'driveable\_surface' & 'highway', 'street', 'roadmarking' \\
            'sidewalk' & 'sidewalk', 'walkway' \\
            'terrain' & 'turf', 'grass', 'sand' \\
            'manmade' & 'building', 'wall', 'fence', 'pole', 'sign', 'light', 'bridge', 'billboard' \\
            'vegetation' & 'bush', 'plants', 'tree' \\
            \hline
        \end{tabular}
    \end{center}
\end{table*}

\end{document}